\documentclass{article}
\usepackage{amssymb}
\usepackage{verbatim}
\usepackage{graphicx}
\usepackage{amsmath}
\usepackage{wrapfig}
\usepackage[preprint]{corl_2025} % Uncomment for pre-prints (e.g., arxiv); This is like ``final'', but will remove the CORL footnote.

\usepackage{booktabs}  
\usepackage{multirow}   
\usepackage{array}      
\usepackage[table]{xcolor}

\usepackage{amsfonts}
\usepackage{algorithmic}
\usepackage{textcomp}
\usepackage{xcolor}
\usepackage{hyperref}
\usepackage{pifont}
\usepackage{geometry}

\newcommand{\cmark}{\textcolor{green}{\ding{51}}}  % 绿色对勾
\newcommand{\xmark}{\textcolor{red}{\ding{55}}}    % 红色叉

\title{One View, Many Worlds: Single-Image to 3D Object Meets Generative Domain Randomization for One-Shot 6D Pose Estimation}

\author{
Zheng Geng$^{*, 1}$, Nan Wang$^{1}$, Shaocong Xu$^{1}$\vspace{0.1cm},\\
\textbf{Chongjie Ye$^{1,5}$, Bohan Li$^{6,7}$, }
\textbf{Zhaoxi Chen$^4$, Sida Peng$^2$, Hao Zhao\footnotemark[2] $^{\ ,1,3}$}\vspace{0.15cm}\\
    $^1$ Beijing Academy of Artificial Intelligence, BAAI;
    $^2$Zhejiang University\\
    $^3$ Institute for AI Industry Research (AIR), Tsinghua University\\
    $^4$ Nanyang Technological University;
    $^5$FNii, The Chinese University of Hongkong, Shenzhen\\ 
    $^6$Shanghai Jiao Tong University; 
    $^7$Eastern Institute of Technology, Ningbo \\
    \vspace{0.15cm}\\
\small
 \texttt{zhenggeng@bit.edu.cn}$^*\quad$
 \texttt{zhaohao@air.tsinghua.edu.cn}$^\dagger$\\
}

\begin{document}
\maketitle

%===============================================================================
\begin{figure}[htbp]
\vspace{-1.2cm} 

\centerline{
\includegraphics[width=0.96\columnwidth]{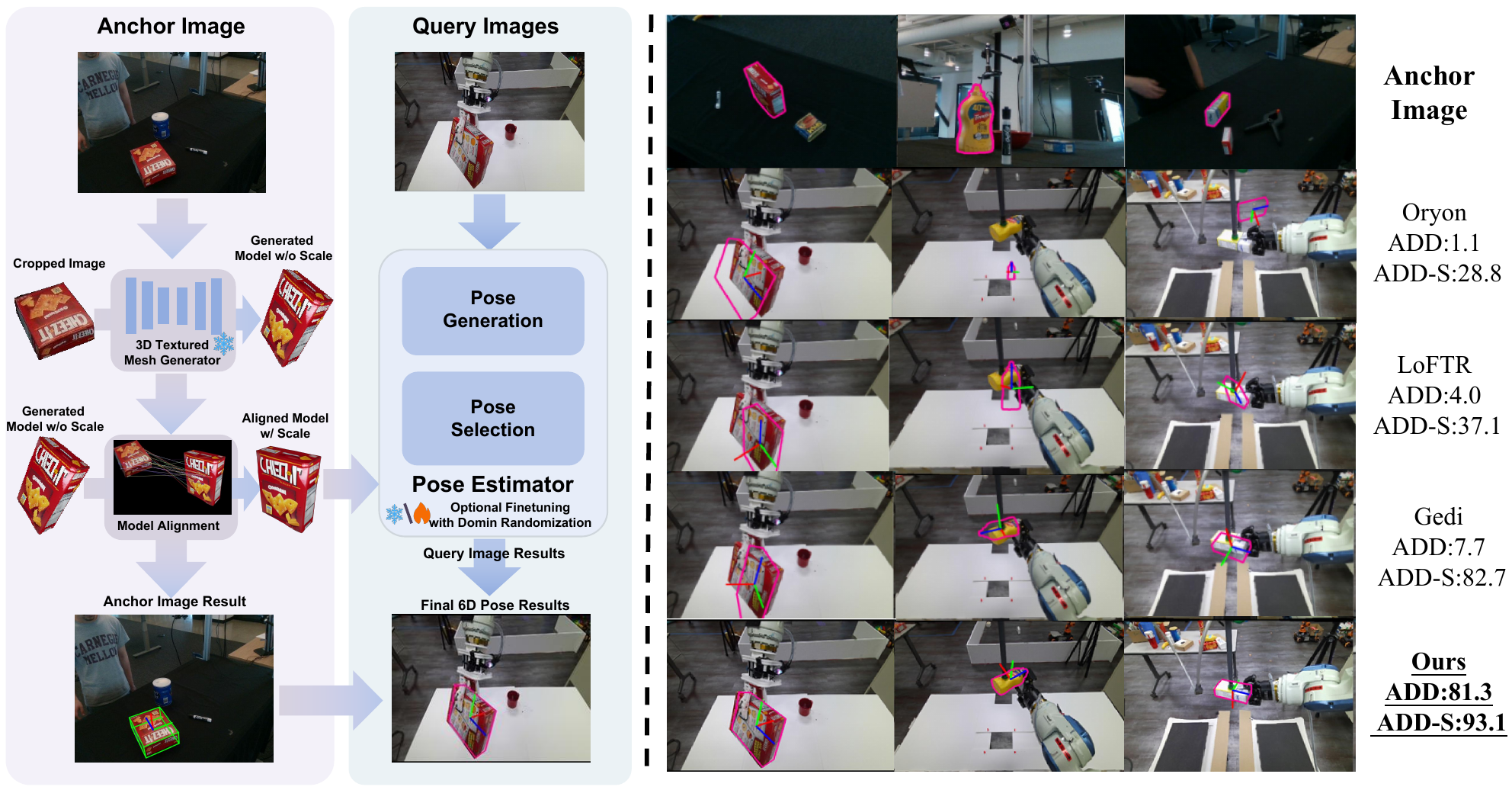}
}

\caption{
\textbf{OnePoseViaGen.} (Left) From a single anchor image, we generate a textured 3D model that lacks real-world scale and pose. Our coarse-to-fine alignment resolves this challenge, and optional domain-randomized fine-tuning further boosts robustness. (Right) While prior methods (Oryon, LoFTR, Gedi) largely fail in this one-shot setting, our approach achieves dramatic gains (ADD 81.3, ADD-S 93.1), enabling reliable 6D pose estimation where baselines break down.
}

\label{fig:teaser}
\vspace{-0.3cm} 
\end{figure}

% \textbf{Hi6DGenPose.} (Left) Our method generates a 3D object model from a single RGB-D anchor image, and recovers the accurate scale and pose of the object through alignment. Using this refined model, we then perform precise 6D pose estimation on the query image using a dedicated pose estimator.
% (Right) Our method demonstrates robust and significant performance improvements compared to existing state-of-the-art approaches, especially under challenging conditions such as occlusion and appearance variation.
% }
\begin{abstract}
Estimating the 6D pose of arbitrary unseen objects from a single reference image is critical for robotics operating in the long-tail of real-world instances. However, this setting is notoriously challenging: 3D models are rarely available, single-view reconstructions lack metric scale, and domain gaps between generated models and real-world images undermine robustness. We propose OnePoseViaGen, a pipeline that tackles these challenges through two key components. \textbf{First}, a coarse-to-fine alignment module jointly refines scale and pose by combining multi-view feature matching with render-and-compare refinement. \textbf{Second}, a text-guided generative domain randomization strategy diversifies textures, enabling effective fine-tuning of pose estimators with synthetic data. Together, these steps allow high-fidelity single-view 3D generation to support reliable one-shot 6D pose estimation. On challenging benchmarks (YCBInEOAT, Toyota-Light, LM-O), OnePoseViaGen achieves state-of-the-art performance far surpassing prior approaches. We further demonstrate robust dexterous grasping with a real robot hand, validating the practicality of our method in real-world manipulation. Project page: \href{https://gzwsama.github.io/OnePoseviaGen.github.io/}{https://gzwsama.github.io/OnePoseviaGen.github.io/}

\end{abstract}

% Two or three meaningful keywords should be added here

%===============================================================================

\section{Introduction}
    Robots operating in the open world must interact with an endless long-tail of novel objects—ranging from an unseen tool on a factory line to an ad-hoc household item in daily life. At the core of this capability lies 6D object pose estimation, the recovery of an object’s precise 3D position $t$ and orientation $R$. However, this task remains notoriously difficult in realistic settings: pre-scanned CAD models are rarely available, multi-view capture is impractical, and single-view reconstructions suffer from scale ambiguity. Consequently, reliable one-shot 6D pose estimation from a single image has long been deemed nearly impossible, despite its central role in simulation~\cite{chen2024urdformer,srinivasan2024dexmots,dwivedi2025interactvlm}, tracking~\cite{zhong2020seeing}, reconstruction~\cite{xia2025drawer,yang2023emernerf,xue2023neural,chen2025hort,rashid2023language,wang2024neural,wu2016single,kulhanek2024wildgaussians,matsuki2024gaussian,dellaert2020shonan,zhou2016fast,choy2019fully,simeonov2023se,zeng20173dmatch,chen2022pq}, and downstream robotic applications such as manipulation~\cite{wang2025articubot,zeng2017,kappler2018real,wen2022catgrasp,blukis2023oneshot,ZHUANG2023robobinpick,eisner2024deep,zhao2025anyplace,mahler2017dex,shen2023distilled,kerr2024robot,caggianomyodex,chen2022system,romero2024eyesight,chen2023visual,bohg2013data,driess2023palm,james2020rlbench,wei2024d,murali2025graspgen,florence2018dense,billard2019trends,guo2017hybrid,deng2019deep,sajjan2020clear,ding2024preafford}, navigation~\cite{gervet2023navigating,forster2014svo}, autonomous control~\cite{kaufmann2018deep,sabbah2025optimal} and teleoperation~\cite{runz2018maskfusion,marchand2015pose}. While VLA models and generalist policies~\cite{kim2024openvla,team2024octo,venkataraman2024real,ahn2022can,zitkovich2023rt,ma2023eureka,o2024open,khazatsky2024droid,kumar2023robohive,chi2023diffusion} excel in broad skills such as folding or bussing, they often fail in precision-critical operations—for example, inserting a charging plug—where sub-centimeter errors lead to task failure. Accurate 6D pose estimation is thus indispensable for reliable robotics.
   
    Most state-of-the-art 6D pose estimation methods are learning-based and can be classified into three categories: instance-level ~\cite{kehl2017ssd6d,peng2019pvnet,wen2020ropose,lim2013parsing}, category-level ~\cite{wang2019densefusion,Chi_2021_ICCV,Li_2020_CVPR,goodwin2022a,li2023sd,lin2021dualposenet}, and category-agnostic ~\cite{nguyen2024gigapose,ornek2024foundpose,wen2024foundationpose,he2022fs6d}. 
    While instance-level methods achieve high accuracy, they are restricted to objects with rich texture in the training set, limiting their use in long tail applications. 
    Category-level methods generalize better within known categories but still struggle with unseen object classes. 
    Recent advances in category-agnostic methods focus on pose estimation for unseen objects using render\&compare strategies~\cite{wen2024foundationpose,labbe2022megapose}, feature matching~\cite{moon2025co,ornek2024foundpose}, or both~\cite{nguyen2024gigapose}, achieving high precision. 
    However, these methods depend on 3D models, which are not trivial to acquire in most scenarios.
    
    To tackle the lack of object models, recent work in model-free pose estimation explores methods that avoid reliance on explicit textured 3D models. These approaches use reference images and have shown promising results, with some employing partial matching for relative pose estimation~\cite{sun2021loftr,corsetti2024open,liu2024unopose} or multiple images to reconstruct the object~\cite{wen2024foundationpose,jin20246dope}. However, many still require multi-view inputs or prior camera-to-object pose knowledge~\cite{liu2024unopose,nguyen2024gigapose}, limiting their effectiveness when only a single image is available but robust 6D pose estimation is needed.
    
    To address these challenges, we propose OnePoseViaGen, a novel model-free method for 6D object pose estimation. Building on single-image 3D generation techniques~\cite{ye2025hi3dgen,guo2025craft,xiang2024trellis}, we extend Hi3DGen~\cite{ye2025hi3dgen} with image-conditioned generation to reconstruct textured 3D object models from a single RGB-D image. Since such reconstructions are scale-normalized, we introduce a coarse-to-fine strategy that jointly refines object scale and 6D pose for accurate alignment. By integrating the refined model with advanced pose estimation frameworks~\cite{wen2024foundationpose}, our method achieves robustness under heavy occlusion, complex lighting, and large viewpoint variations. We evaluate OnePoseViaGen on standard benchmarks (YCBInEOAT, Toyota-Light, LM-O) and a new annotated in-the-wild dataset, and further demonstrate its effectiveness in dexterous grasping tasks. Results show that our method significantly outperforms state-of-the-art approaches for unseen objects. Our contributions are summarized as follows:
    \begin{itemize}
        \item \textbf{Generative pipeline for one-shot 6D pose.} We introduce the first pipeline that integrates single-view 3D generation into both training and inference for one-shot 6D pose and scale estimation, proving that generative modeling can directly benefit pose estimation.

        \item \textbf{Coarse-to-fine metric alignment.} We design a coarse-to-fine alignment module that jointly refines pose and metric scale via 2D–3D feature matching and render-and-compare refinement, enabling accurate metric-scale alignment from a single image.
        
        \item \textbf{Text-guided generative domain randomization.} We propose a text-driven augmentation strategy that generates diverse, structurally consistent 3D variants. Rendering these models under randomized conditions yields a large-scale synthetic dataset, which bridges the domain gap between generated models and real-world objects.
        
        \item \textbf{State-of-the-art performance and real-world validation.} Our method establishes new state-of-the-art results on three challenging 6D pose benchmarks and a newly collected in-the-wild dataset. We further demonstrate its effectiveness in dexterous robotic grasping, validating robustness and practicality in real-world manipulation. 
    \end{itemize}

%===============================================================================

\section{Related Works}
\label{sec:Related work}

\subsection{6D Pose Estimation}  
The field of 6D pose estimation has been extensively explored through both model-based and model-free approaches, each addressing the challenge of localizing objects in 3D space with varying strategies~\cite{Lim_2013_ICCV,Wang_2019_CVPR,liu2022gen6d,sun2022onepose,labbe2022megapose,Huang2023,ornek2024foundpose,nguyen2024gigapose,wen2024foundationpose}. Model-based methods typically rely on CAD models or 3D representations for pose estimation. For instance, MegaPose ~\cite{labbe2022megapose} employs a render\&compare strategy to refine poses of novel objects using synthetic views of CAD models. Similarly, FoundPose ~\cite{ornek2024foundpose} leverages foundation features~\cite{zhong20233d,zhong2022snake,oquab2023dinov2} from self-supervised vision models to establish 2D-3D correspondences between RGB images and pre-rendered templates, demonstrating robust generalization without task-specific training. FS6D ~\cite{he2022fs6d} extends this domain by addressing few-shot pose estimation, introducing a dense prototype matching framework and a large-scale photorealistic dataset (ShapeNet6D) to enhance generalization. In contrast, model-free approaches aim to reduce reliance on explicit 3D models. Wang et al. ~\cite{Wang_2019_CVPR}designed loss functions for specific symmetry types. OnePose ~\cite{sun2022onepose} and its successor, OnePose++ ~\cite{he2022onepose++}, eliminate the need for CAD models by reconstructing sparse or semi-dense object models from RGB videos, enabling real-time pose estimation even for low-textured objects. GigaPose ~\cite{nguyen2024gigapose} achieves fast and accurate pose estimation by leveraging discriminative templates and patch correspondences, offering significant speedups while maintaining robustness to segmentation errors. FoundationPose ~\cite{wen2024foundationpose} further advances the field by proposing a unified framework that supports both model-based and model-free setups, utilizing neural implicit representations for novel view synthesis when CAD models are unavailable. Gen6D ~\cite{liu2022gen6d} introduces a generalizable model-free approach that predicts poses of unseen objects using only posed RGB images. 
Despite these advancements, existing methods are limited to textured objects and require costly retraining for novel instances. In this work, we propose a novel model-free approach that leverages textured 3D model generation to achieve accurate pose estimation without prior 3D data.\looseness=-1

\subsection{3D Model Generation}  

The generation of high-fidelity 3D models from 2D inputs has emerged as a critical area of research~\cite{stoiber2022srt3d,
xiang2024structured,chen2024rapid,chen2024vp3d,chen2024it3d,tang2025cycle3d,xie2024styletex,wu2024consistent3d,xu20243d,ye2025hi3dgen}, with significant implications for applications such as 6D pose estimation. Hi3DGen ~\cite{ye2025hi3dgen} introduces a novel framework that leverages normal maps as an intermediate representation to bridge the gap between 2D images and detailed 3D geometry. By decoupling low- and high-frequency image patterns through noise injection and dual-stream training, Hi3DGen achieves superior fidelity in reproducing fine-grained geometric details, outperforming existing methods in generating high-quality 3D models. Similarly, TRELLIS ~\cite{xiang2024structured} presents a scalable approach to 3D generation by integrating sparse 3D grids with dense multiview visual features, enabling versatile decoding into formats like Radiance Fields, 3D Gaussians, and meshes.  Rapid 3D Model Generation highlights the potential of AR/VR devices as intuitive tools for 3D modeling, introducing Deep3DVRSketch~\cite{chen2024rapid}, which enables novice users to create detailed 3D models from simple sketches efficiently. VP3D~\cite{chen2024vp3d} enhances text-to-3D generation by incorporating 2D visual prompts, improving visual fidelity and texture detail through explicit visual appearance guidance. IT3D~\cite{chen2024it3d} further refines this process by synthesizing multi-view images explicitly, tackling issues like over-saturation and inadequate detailing via a Diffusion-GAN dual training strategy. Cycle3D~\cite{tang2025cycle3d} presents a unified framework that cyclically utilizes 2D diffusion and 3D reconstruction modules to ensure high-quality and consistent 3D content generation. StyleTex~\cite{xie2024styletex} focuses on style-guided texture generation for 3D models, employing a diffusion-model-based approach to harmonize textures with reference images and text descriptions. Consistent3D~\cite{wu2024consistent3d} addresses inconsistencies in text-to-3D generation by exploring deterministic sampling priors through ODE trajectory sampling, ensuring more reliable and high-fidelity outputs. 
However, these methods generate normalized models with scale mismatches to real-world objects, leading to errors in downstream pose estimation. Our OnePoseViaGen introduces a coarse-to-fine strategy that jointly refines the scale and 6D pose of generated models, enabling accurate recovery of real-world object dimensions and scene poses.\looseness=-1

%===============================================================================

\section{Method}
\label{sec:Method}
\subsection{Problem Formulation}
\begin{figure}[htbp]
\vspace{-0.2cm} 
\centerline{
\includegraphics[width=0.85\columnwidth]{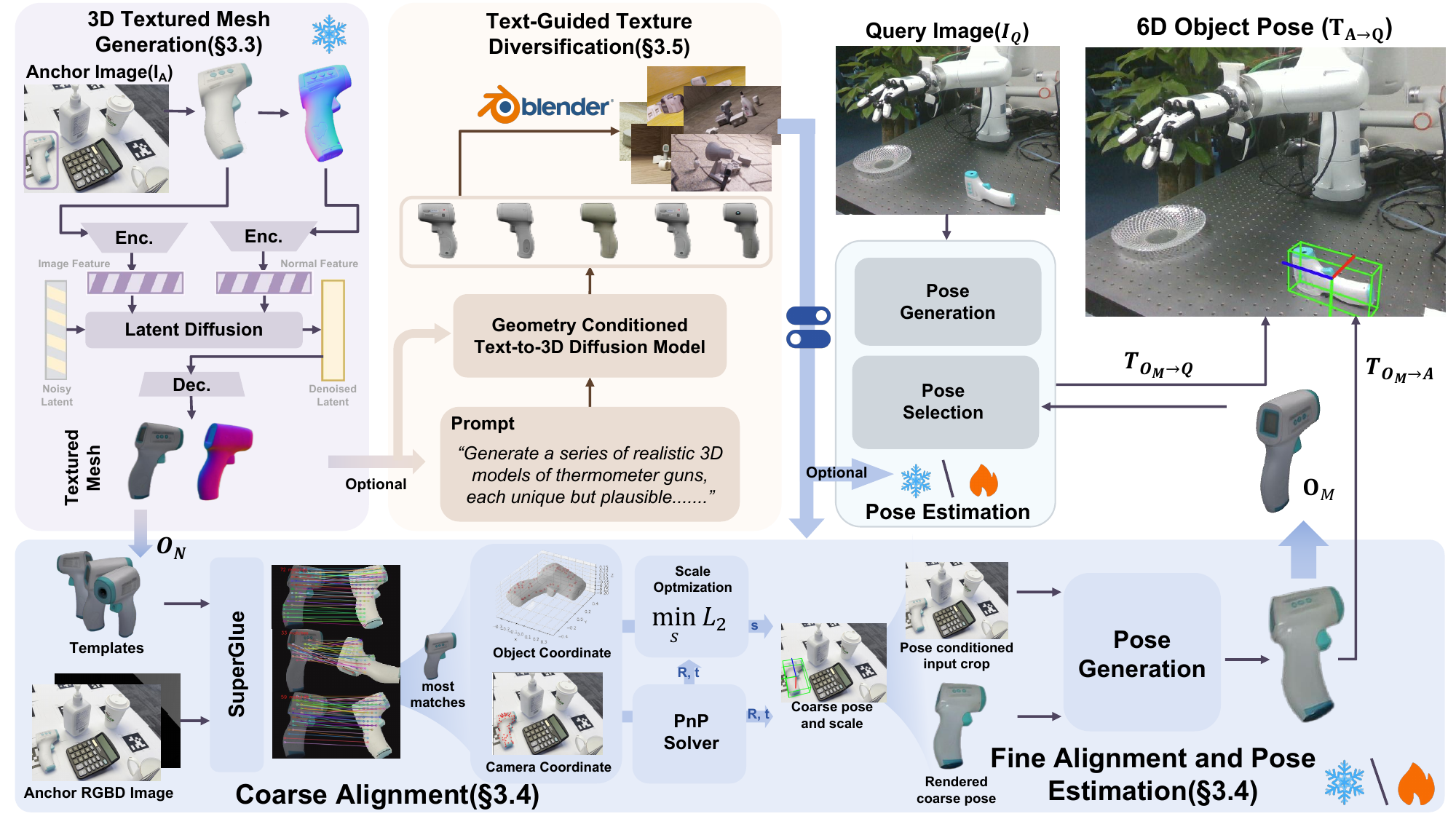}
}
\vspace{-0.1 cm} 
\caption{
\textbf{Overview of OnePoseViaGen.} Given an anchor RGB-D image $\mathbf{I}_A$, we generate a textured 3D model $O_N$ via our normal-guided generative pipeline. To ground $O_N$ in real-world metrics, we align it with $\mathbf{I}_A$ via multi-view feature matching and render\&compare framework. For a query image $\mathbf{I}_Q$, we apply a render\&compare\&selection pipeline~\cite{wen2024foundationpose}. Finally, the relative transformation $\mathbf{T}_{A \rightarrow Q}$ is computed. To enhance generalization, we introduce a text-prompt-based diversification strategy for training data, bridging the domain gap between generated models and real objects.
}

\label{fig:overview}
\vspace{-0.6 cm} 
\end{figure}

    As illustrated in Fig.~\ref{fig:overview}, let $\mathbf{I}_A$ (top-left corner) and $\mathbf{I}_Q$ (top-right) denote an RGB-D anchor image and a query image, respectively. Our task is to estimate the relative rigid transformation $\mathbf{T}_{A \rightarrow Q} \in SE(3)$ that maps the pose of a target object from the anchor image coordinate system to that of the query image. Since generated models are scale-normalized, we additionally determine a scaling factor $s$ based on $\mathbf{I}_A$ to calibrate the model to real-world scale. Formally, $\mathbf{T}_{A \rightarrow Q} = [\mathbf{R} \mid \mathbf{t}]$. The core challenge lies in achieving robust pose estimation when the query image may capture the same object under significantly different conditions. Our method thus needs to extract invariant geometric features and establish reliable correspondences despite such appearance changes.
\subsection{Approach Overview}
    Fig.~\ref{fig:overview} outlines our OnePoseViaGen pipeline. Given an RGB-D anchor image $\mathbf{I}_A$ (top-left of Fig.~\ref{fig:overview}) of a novel object—for which no pre-existing 3D model exists—our key challenge is to enable 6D pose estimation from a single view. To address this, we first generate a textured 3D model with standardized orientation/scale from $\mathbf{I}_A$ using Hi3DGen\cite{ye2025hi3dgen} in single-view 3D generation (Sec.~\ref{sec:generation}). Critical to grounding this model in the real world, our coarse-to-fine alignment module (Sec.~\ref{sec:coarsealign}) jointly recovers the object’s metric scale and 6D pose in $\mathbf{I}_A$.   
    With this metric-calibrated model, we estimate the object’s pose in any query RGB-D image $\mathbf{I}_Q$ (top-right of Fig.~\ref{fig:overview}) using the aligned model and a robust pose estimation framework. The final relative transformation $\mathbf{T}_{A \rightarrow Q}$ is then computed from the absolute poses in both views. 
    
    To bridge the domain gap between generated models and real-world images, we introduce a text-guided generative augmentation strategy (Sec.~\ref{sec:finetune}). This generates structurally consistent but texture-diverse 3D variants, which are rendered under randomized conditions (lighting, backgrounds, occlusions) to create a large synthetic dataset. This dataset enables fine-tuning of pose estimation components, significantly boosting robustness—validated in our experiments (Sec.~\ref{sec:result}).
    
\subsection{Normal-guided 3D Textured Mesh Generation}
    \label{sec:generation}
    \paragraph{Motivation.} Estimating the 6D pose of novel objects typically requires a 3D model, which is often unavailable. While multi-view reconstruction can generate models from multiple images with known poses~\cite{wen2024foundationpose,he2022onepose++,he2022fs6d,liu2022gen6d}, this is not feasible in our one-shot setting. 
    Neural implicit representations can also generate 3D geometry but often require time-consuming optimization for each object~\cite{mueller2022instant,wang2021neus,wen2023bundlesdf}.
    To overcome these limitations and enable pose estimation from a single view, our first step is to generate a textured 3D object model directly from the input anchor image $I_A$. Leveraging recent advances in single-view generative 3D modeling allows us to quickly obtain a plausible 3D representation without the need for multi-view data or lengthy per-object optimization.
    
    \paragraph{Standardized Textured Mesh Geneartion with Normal and Image conditions.} As depicted in the ``3D Textured Mesh Generation" step in Fig.~\ref{fig:overview}, we generate high-fidelity textured meshes from a single image based on a modified version of Hi3DGen~\cite{ye2025hi3dgen}.
    Initially, segmentation is used to crop the input anchor image to minimize background noise's impact on the model generation process. 
    Subsequently, the processed image $I_A^{\text{cropped}}$ undergoes an off-the-shelf image-to-normal estimation to produce a normal map $X$. The normal maps $X$ and the original cropped image $I_A^{\text{cropped}}$ are then fed into the 3D generation model. 
    This process ensures geometric consistency and visual detail. 

    Finally, we obtain a standardized textured object model $O_N$ in an object-centric coordinate system. This model $O_N$ serves as the basis for subsequent alignment step to match real-world dimensions, as outlined in Sec.~\ref{sec:coarsealign}. We further visualize several examples of these generated models in Appendix.~\ref{additional_exp}, demonstrating the fidelity and diversity achieved.

\subsection{Coarse to Fine Object Alignment and Pose Estimation}
\label{sec:coarsealign}
    \begin{wrapfigure}{r}{0.65\columnwidth}
    \centering
    \includegraphics[width=0.65\columnwidth]{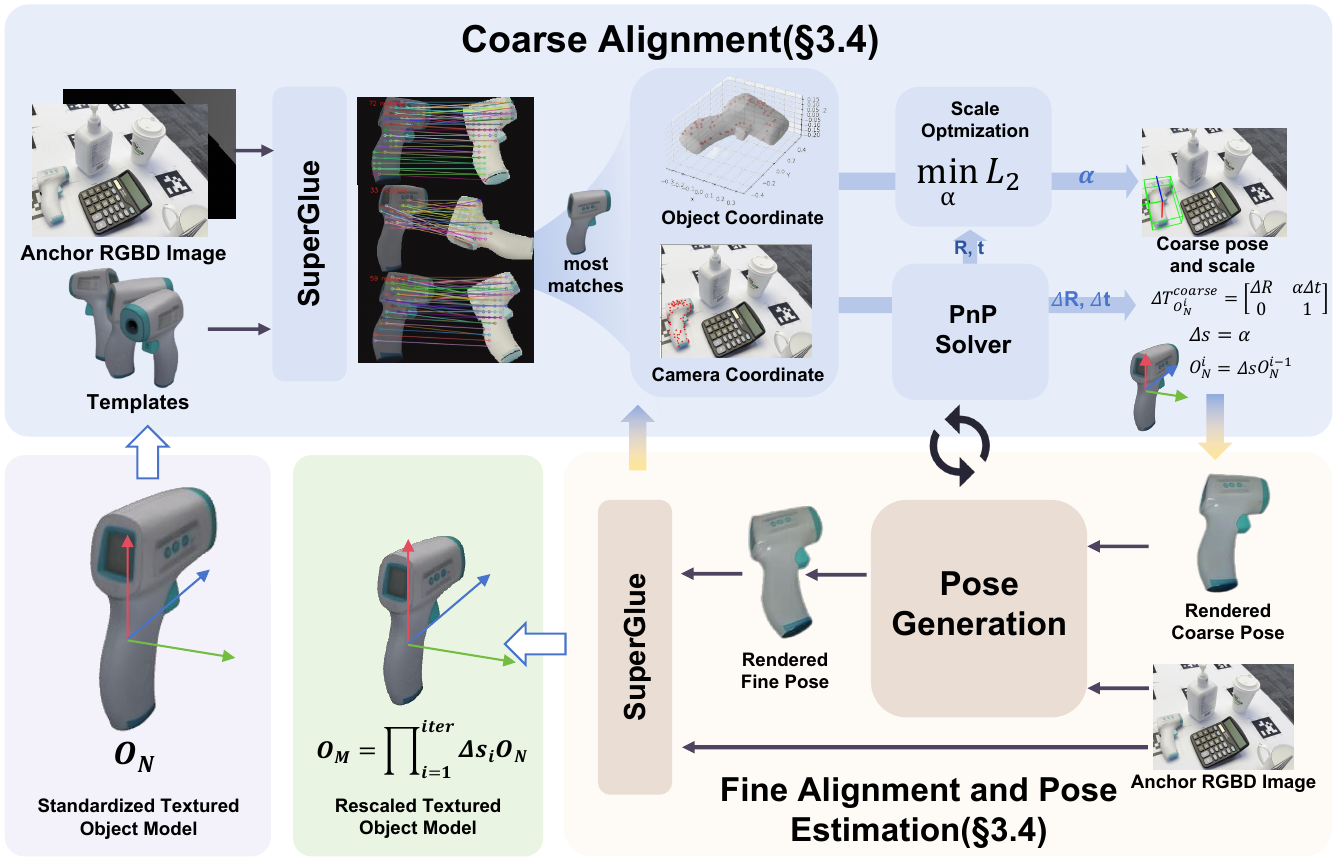}
    \caption{\textbf{Overview of Coarse-to-fine Alignment Process.}}
    \label{fig:alignment_detail}
\end{wrapfigure}
    \paragraph{Motivation.} Although Sec.~\ref{sec:generation} provides a 3D shape $O_N$, it is in a normalized space, not matching the real object's scale and pose in $\mathbf{I}_A$. For accurate metric-space 6D pose estimation, determining the correct scale and aligning the model with the observed object in the 3D scene is essential. This alignment is challenging due to discrepancies between the idealized model and noisy, partial observations from $\mathbf{I}_A$. To address this, we propose a multi-stage coarse-to-fine alignment strategy, illustrated in Fig.~\ref{fig:alignment_detail}.

    \paragraph{Coarse Alignment via Multi-view Feature Matching and PnP solving.}
    In coarse stage, we render object $O_N$ from $n$ spherical viewpoints using camera intrinsics $\mathbf{K}$. SuperPoint~\cite{detone2018superpoint} features are extracted and matched via SuperGlue~\cite{Sarlin2020superglue} between each rendered view $o^i$ and $\mathbf{I}_A$, and the view with the most matches is selected. Matched 2D point pairs $(p'_i \in \mathbb{R}^2, p_i \in \mathbb{R}^2)$ (from the rendered view and $\mathbf{I}_A$, respectively) are lifted to 3D points $(\mathbf{P'_i} \in \mathbb{R}^3, \mathbf{P_i} \in \mathbb{R}^3)$ in the object's normalized coordinate system and the camera coordinate system using depth information from the rendered view and
    $\mathbf{I}_A$:
    % in the object coordinate system and the camera coordinate system, respectively, according to:
    $
    \mathbf{P'}_i^{N} = \frac{1}{d'_i} \cdot \mathbf{K}^{-1} \cdot [\mathbf{p'}_i^{N}, 1]^\top, \quad
    \mathbf{P}_i^{N} = \frac{1}{d_i} \cdot \mathbf{K}^{-1} \cdot [\mathbf{p}_i^{N}, 1]^\top
    $. Subsequently, we apply the Perspective-n-Point (\textbf{PnP}) algorithm~\cite{MartinPnP1987} on $\textbf{P}'_i$ and their corresponding 2D projections $\textbf{p}_i$ in image $o^i$ to estimate an initial 6-DoF pose, with scale ambiguity.
    
    Then, we transform normalized model point $ \mathbf{P'_i} $ into the same camera coordinate system:
    $
        \mathbf{\hat{P}_i} = R \mathbf{P'_i} + t.
    $
    Ideally, $ \mathbf{\hat{P}_i} $ and corresponding camera-space points $ \mathbf{P}_i $ represent the same 3D point, differing only by a global scale factor $ \alpha > 0 $, where $ \alpha $ denotes the scaling factor from the origin of the camera coordinate system along the viewing direction:
    $
        \mathbf{P}_i^{N} = \alpha \mathbf{\hat{P}_i}.
    $
    
    To estimate $ \alpha $, we minimize the following least-squares objective:
    $
        L_2 = \sum_i \left\| \alpha \mathbf{\hat{P}_i} - \mathbf{P}_i^{N} \right\|^2.
    $
    This leads to a single-variable optimization problem: 
    $
        \min_\alpha L_2,
    $
    which yields the optimal $ \alpha $. And due to the scale consistency property of similarity transformations, the final model's scale remains invariant to the choice of scaling center. This $ \alpha $ also serves as the desired scale factor $ s $, defined with respect to the model center as the origin. Based on this optimization, we obtain a coarse estimate of the object's pose 
    \[
    \mathbf{T}_{O_N \to A}^{\text{coarse}}=\begin{bmatrix}
        R & \alpha t \\
        \mathbf{0} & 1
    \end{bmatrix}.
    \]
    
    and scaling factor $s=\alpha$. Since the process is iterative, we also express the transformation incrementally: 
    \[
    \Delta \mathbf{T}_{O_N \to A}^{\text{coarse}} = 
    \begin{bmatrix}
        \Delta R & \alpha \Delta t \\
        \mathbf{0} & 1
    \end{bmatrix}, \quad
    \Delta s = \alpha.
    \]
    \paragraph{Fine Alignment and Pose Estimation via Render-and-Compare Refinement.}
    The coarse alignment provides a reasonable initial pose and scale, but inaccuracies due to feature noise and model imperfections necessitate further refinement. We iteratively refine the pose and scale using a strategy that combines render\&compare refinement with the coarse alignment's scale optimization. Starting with the coarse estimate $\mathbf{T}_{O_N \to A}^{\text{coarse}}$ and scale $s$, we refine the estimation using the network adapted from FoundationPose~\cite{wen2024foundationpose} to predict incremental pose updates $\Delta R \in SO(3)$ and $\Delta t$ based on the difference between a rendering of the model at the current pose and the observed object in $\mathbf{I}_A$. The pose is updated as
     $
        t^+ = t + \Delta t,~~
        R^+ = \Delta R \otimes R,
    $
    where $\otimes$ denotes update on SO(3). Crucially, after each pose update, we re-run the feature matching and scale optimization from the coarse stage using the current pose estimation to refine the scale $s$. This iterative process, alternating between pose refinement and scale optimization, continues until convergence or a maximum number of iterations is reached. This yields a metric-scale 3D model $O_M =  {\textstyle \prod_{i}^{iter}} \Delta s_iO_N$ and its accurate 6D pose $\mathbf{T}_{O_M \to A}$ relative to the anchor image frame $\mathbf{I}_A$. The effectiveness of this coarse-to-fine approach, particularly the fine alignment stage, is demonstrated in the ablation study (Sec.~\ref{ablation}).

    For estimating the poses in query images $I_Q$ (illustrated in Fig.~\ref{fig:overview}), we utilize the derived metric-scale model $O_M$. We employ a render\&compare\&selection strategy from FoundationPose ~\cite{wen2024foundationpose}. The resulting 6D pose relative to the camera frame in query image $I_Q$ is denoted as $\mathbf{T}_{O_M \to Q}$. Finally, as shown in Fig.~\ref{fig:overview}, the relative transformation between the anchor and query views, is computed using the estimated absolute poses: 
    \begin{equation}
        \label{eq:relative-pose}
        \mathbf{T}_{A \to Q} = (\mathbf{T}_{O_M \to A})^{-1} \cdot \mathbf{T}_{O_M \to Q}.
    \end{equation}
    The quantitative results showcasing the accuracy of our 6D pose estimation are presented in Table~\ref{tab:ycbineoat}, Table~\ref{tab:toyota-light}, and Table~\ref{tab:lmo} in Sec.~\ref{sec:result}.

\subsection{Text-Guided Texture Diversification}
\label{sec:finetune}
    \paragraph{Motivation.} 
    The 3D model $O_N$ generated in Sec.~\ref{sec:generation} captures shape and initial texture, but represents only a single instance. It lacks variation in appearance and geometry found in real-world objects, limiting its generalization under diverse environmental conditions (e.g., lighting, occlusion). Training robust pose estimators requires large, diverse datasets, which are costly to collect manually. To bridge the gap between generated models and real-world objects and facilitate effective training, we propose generating a diverse set of plausible object variations.
    
    \paragraph{Texture Diversification with Structure-Aware Text-to-3D.}
    As illustrated in the "Generative Domain Randomization" step in Fig.~\ref{fig:overview}, we use the text-guided 3D generation model Trellis~\cite{xiang2024trellis}, taking the initial model $O_N$ and text prompts as input (Appendix.~\ref{diversity}). The model generates variants with diverse textures, styles, or minor geometric changes while preserving core structure. We render these models under randomized viewpoints, lighting and occlusions using Blender, forming a large-scale synthetic dataset for fine-tuning the render-and-compare network (Sec.~\ref{sec:coarsealign}). The method improves the pose generation model on the long-tail category through diverse texture synthesis. Incorporating category-level training data enables the pose estimation module to significantly improve its performance—particularly on objects that were previously challenging, or when the occluded side appears in query images. As a result, the overall generalization capability of the pose estimation module is notably enhanced. As shown in Table~\ref{tab:ablation_table}, this significantly improves pose estimation performance.
	
%===============================================================================
\section{Experimental Results}
\label{sec:result}

\begin{table*}[h]
\vspace{-0.7cm}
\caption{ Comparison with SOTA on the YCBInEOAT Dataset. }

    \centering
    \resizebox{1\columnwidth}{!}{
    
    \begin{tabular}{llccccccc>{\columncolor{gray!20}}c>{\columncolor{gray!20}}c}
        \toprule
        ~ & \multicolumn{2}{c}{Oryon~\cite{corsetti2024open}} & \multicolumn{2}{c}{LoFTR~\cite{sun2021loftr}} & \multicolumn{2}{c}{Gedi~\cite{poiesi2022learning}} & \multicolumn{2}{c}{Any6D~\cite{lee2025any6d}} & \multicolumn{2}{c}{\cellcolor{gray!20}Ours} \\ 
        \cmidrule(lr){2-3} \cmidrule(lr){4-5} \cmidrule(lr){6-7} \cmidrule(lr){8-9} \cmidrule(lr){10-11}
        Modality & \multicolumn{2}{c}{RGB-D \& Language} & \multicolumn{2}{c}{RGB-D}& \multicolumn{2}{c}{Depth} & \multicolumn{2}{c}{RGB-D}& \multicolumn{2}{c}{\cellcolor{gray!20}RGB-D}\\

        Metrics & ADD-S($\uparrow$) & ADD($\uparrow$) & ADD-S($\uparrow$) & ADD($\uparrow$) & ADD-S($\uparrow$) & ADD($\uparrow$) & ADD-S($\uparrow$) & ADD($\uparrow$) & ADD-S($\uparrow$) & ADD($\uparrow$)  \\ 

        \midrule
        
        sugar\_box1 & 44.0 & 0.0 & 47.3 & 0.0 & 95.6 & 0.0 & 96.7 & 14.3 & 93.71 & 75.63  \\ 
        sugar\_box\_yalehand0 & 34.7 & 3.0 & 41.6 & 0.0 & 82.2 & 6.9 & 89.1 & 75.2 & 94.59 & 87.12  \\

        mustard0 & 48.6 & 0.0 & 47.3 & 20.3 & 100 & 0.0 & 100 & 23 & 96.98 & 70.00  \\ 
        mustard\_easy\_00\_02 & 36.2 & 0.0 & 23.2 & 0.0 & 78.3 & 0.0 & 78.3 & 53.6 & 85.58 & 70.12  \\

        bleach0 & 10.4 & 0.0 & 55.2 & 0.0 & 74.6 & 0.0 & 98.5 & 68.7 & 95.84 & 92.41  \\ 
        bleach hard\_00\_03\_chaitanya & 24.4 & 6.7 & 60 & 15.6 & 66.7 & 62.2 & 73.3 & 51.1 & 94.39 & 91.35  \\

        tomato\_soup\_can\_yalehand0 & 32.8 & 0.0 & 10.7 & 0.0 & 60.3 & 0.0 & 70.2 & 0.0 & 88.77 & 77.72  \\

        cracker\_box\_reorient & 13.2 & 0.0 & 26.3 & 0.0 & 97.4 & 0.0 & 100 & 60.5 & 96.34 & 79.40  \\

        cracker\_box\_yalehand0 & 15.0 & 0.0 & 22.6 & 0.0 & 89.5 & 0.0 & 97.7 & 63.9 & 91.64 & 87.64  \\ 

        \midrule
        MEAN & 28.8 & 1.1 & 37.1 & 4.0 & 82.7 & 7.7 & 89.3 & 45.6 & \textbf{93.10} & \textbf{81.27} \\ 
        
    \bottomrule
    \end{tabular}
    \label{tab:ycbineoat}
    }
    \vspace{-0.1cm}
\end{table*}

\begin{table}[h]
\vspace{-0.3cm}
\tiny
\centering
 \begin{minipage}[t]{0.45\textwidth}  
\caption{Comparison on the TOYL Dataset.}
    \centering
    \begin{tabular}{lcccc}
        \toprule
        Methods & AR($\uparrow$) & MSSD($\uparrow$) & MSPD($\uparrow$) & VSD($\uparrow$)  \\ 
        \midrule
        SIFT\cite{lowe1999object} & 30.3 & 39.6 & 44.1 & 7.3  \\ 
        Obj. Mat.~\cite{gumeli2023objectmatch} & 9.8 & 13.0 & 14.0 & 2.4  \\ 
        Oryon~\cite{corsetti2024open} & 34.1 & 42.9 & 45.5 & 13.9  \\ 
        Any6D~\cite{lee2025any6d} & 43.3 & 55.8 & 58.4 & 15.8  \\ 
        \rowcolor{gray!20} Ours & \textbf{55.7} & \textbf{67.0} & \textbf{65.1} & \textbf{35.1} \\ 
    \bottomrule
    \end{tabular}
    \label{tab:toyota-light}
\end{minipage}  
\hspace{0.05\textwidth} 
\begin{minipage}[t]{0.45\textwidth}  
\caption{Comparison on the LINEMOD Occlusion (LM-O) Dataset.}
    \centering
    \begin{tabular}{lcccc}
        \toprule
        Methods & AR($\uparrow$) & MSSD($\uparrow$) & MSPD($\uparrow$) & VSD($\uparrow$)  \\ 
        \midrule
        GigaPose~\cite{nguyen2024gigapose} & 17.5 & 35.8 & 9.0 & 7.6  \\ 
        Any6D~\cite{lee2025any6d} & 28.6 & 36.1 & 32.0 & 17.6  \\ 
        \rowcolor{gray!20} Ours & \textbf{74.8} & \textbf{79.5} & \textbf{84.9} & \textbf{60.0} \\ 
    \bottomrule
    \end{tabular}
    \label{tab:lmo}
\end{minipage}  
\vspace{-0.4cm}
\end{table}

\subsection{Datasets and Setup}

\textbf{Public datasets.}~~We evaluated our method without the proposed fine-tuning (Sec.~\ref{sec:finetune}) to ensure fairness on three public datasets: YCBInEOAT~\cite{wen2021data} (robotic interaction), Toyota-Light (TOYL)~\cite{hodan2018bop} (challenging lighting), and LINEMOD Occlusion (LM-O)~\cite{brachmann2014learning} (occluded, textureless).

\textbf{Real-world evaluation.}~~We performed two experiments in real-world settings: (1) 6D pose estimation for uncommon objects by generating synthetic training data via our domain randomization pipeline and testing on a calibrated real set (built based on~\cite{hodan2018bop}), and (2) robotic manipulation tasks, establishing grasping setups using a ROKAE robot arm equipped with an XHAND1 dexterous hand, and two AgileX PiPERs, and measuring success rates against baselines.

% \subsection{Metrics}

% We adopted evaluation protocols based on the following standard metrics:
% \begin{enumerate}
%     \item The area under the curve (AUC) of ADD and ADD-S metrics. ADD computes the mean distance between corresponding points on the predicted 3D model and ground truth model. For objects with symmetries, ADD-S calculates the average distance between points on the predicted model and their closest corresponding points on the ground truth model.
%     \item The average recall (AR) across the VSD, MSSD and MSPD metrics, as established in the BOP challege benchmark~\cite{hodan2018bop}.
% \end{enumerate}
\subsection{Comparison with the State of the Art}
\textbf{Quantitative results.}~~Following the evaluation settings of Any6D~\cite{lee2025any6d}, we align the relative pose estimate $\mathbf{T}_{A\rightarrow Q}$ according to Eq.~\eqref{eq:relative-pose}. Our approach is primarily compared against recent state-of-the-art methods, including Any6D~\cite{lee2025any6d}, Gedi~\cite{poiesi2022learning}, LoFTR~\cite{sun2021loftr}, and Oryon~\cite{corsetti2024open}.
On YCBInEOAT, anchor images are sampled from DexYCB~\cite{chao2021dexycb}. As shown in Table~\ref{tab:ycbineoat}, our method achieves significantly better robustness. Notably, on challenging objects such as \{\text{sugar\_box1}, \text{mustard0}, \text{tomato\_soup\_can\_yalehand0}\}, previous methods achieved very low ADD scores (e.g., Any6D: $14.3\%$, $23\%$, $0.0\%$), while our method consistently performs well ($75.63\%$, $70.00\%$, $77.72\%$), achieving a mean ADD of $81.27$.
For TOYL (Table~\ref{tab:toyota-light}) and LM-O (Table~\ref{tab:lmo}), we report BOP benchmark metrics. Our method shows consistent improvements across all evaluation criteria.

\subsection{Experimenting in Robotic Manipulation}

% \begin{figure}[htbp]
\begin{wrapfigure}{r}{0.55\columnwidth}
% \vspace{-0.6cm}
\centerline{
\includegraphics[width=0.55\columnwidth]{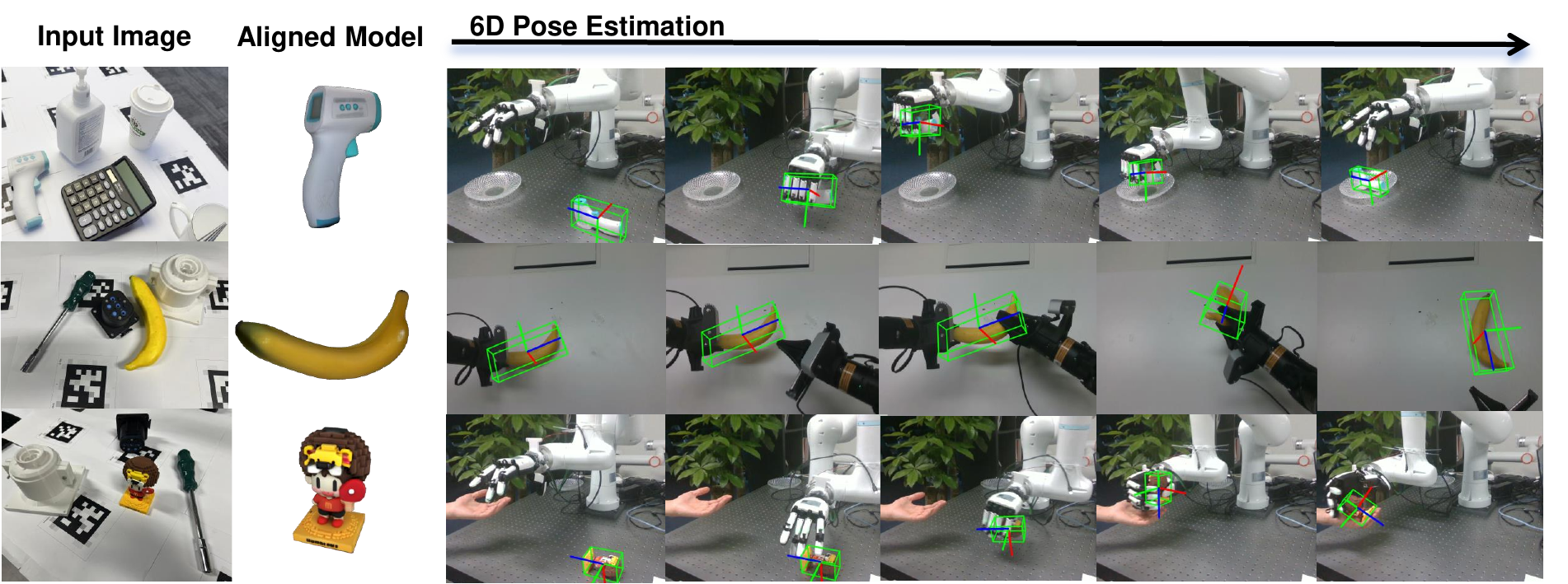}
}

\caption{
\textbf{Qualitative results of robot manipulation.} The left column is the anchor images, the middle column shows the generated models, and the right part shows the pose estimation results during robot manipulation.
}

\label{fig:robot-visual}
% \vspace{-0.6cm}
% \end{figure}
\end{wrapfigure}
To validate the practical applicability and robustness of our method, we conducted two real-world robotic manipulation tasks using 6D pose estimates as the perception input (See Appendix.\ref{additional_exp} for details). 
(1) Pick-and-place: A dexterous hand mounted on a robot arm performs grasping. This task is challenging due to heavy occlusions during grasp.
(2) Dual-arm manipulation: (pick, handoff, place): This involves coordinated control, object instability during transfer, and dynamic occlusions.
We tested both tasks on 15 objects over 30 trials each, comparing our method with SRT3D~\cite{stoiber2022srt3d} and DeepAC~\cite{wang2023deep}. As shown in Fig.~\ref{fig:robot-visual}, our method achieved a success rate of $73.3\%$ (Table~\ref{tab:in-the-wild}), demonstrating strong performance in complex, real-world scenarios.
% \input{tables/ablation_table}

% \begin{table}[t]
% \tiny
% \centering
% \begin{minipage}[t]{0.40\textwidth} 
% \caption{Training strategy effectiveness.}
%     \centering
%     \begin{tabular}{lc}
%       \toprule
%         Methods & AR($\uparrow$)\\ 
%         \midrule
%         Baseline & 12.6  \\ 
%         \textit{w/ naive finetuning} & 11.4 \\ 
%         {\rowcolor{gray!20} \textit{w/ new finetuning} & 52.4 }\\ 
%     \bottomrule
%     \end{tabular}
%     \label{tab:training-strategy}
% \end{minipage}
% \hspace{0.05\textwidth}
% \begin{minipage}[t]{0.40\textwidth} 
% \caption{ Ablation Study.}
%     \centering
%     \begin{tabular}{lcccc}
%         \toprule
%         Method & AR($\uparrow$) & CD($\downarrow$)\\
%          \midrule
%          RosePose & 55.7 & 0.72 \\ 
%          \textit{w/o coarse align} & 54.2 & 0.94\\ 
%          \textit{w/o fine align} & 32.9 & 1.05\\ 
%         \textit{w/o align} & 0.0 & 38.47 \\
%     \bottomrule
%     \end{tabular}
%     \label{tab:ablation}
% \end{minipage}
% \vspace{-0.6cm}
% \end{table}

% \hspace{0.1\textwidth}
% \begin{minipage}[tc]{1\textwidth} 
% \caption{ Real-world evaluation.}
%     \centering
%     \begin{tabular}{lcccc}
%         \toprule
%         Methods & SRT3D & DeepAC & {\cellcolor{gray!20} Ours} \\ 
%         \midrule
%         Success Rate & 6.7 & 16.7 & {\rowcolor{gray!20}73.3} \\ 
%     \bottomrule
%     \end{tabular}
% \end{minipage}

\begin{table}[htbp]
\vspace{-0.2cm}
\tiny
\centering
% \hspace{0.05\textwidth}
\begin{minipage}[t]{0.65\textwidth} 
\caption{ Ablation Study. }

    \centering
    \resizebox{1\columnwidth}{!}{
    
    \begin{tabular}{lccccccc}
        \toprule
        ~Dataset & \multicolumn{4}{c}{TYOL Dataset~\cite{hodan2018bop}} & \multicolumn{3}{c}{Annotated Real-World Dataset} \\ 
        \cmidrule(lr){2-5} \cmidrule(lr){6-8} 
        Method & \multicolumn{1}{c}{\shortstack{w/ \\ align}} & \multicolumn{1}{c}{\shortstack{w/o \\ coarse align}} & \multicolumn{1}{c}{\shortstack{w/o \\ fine align}} & \multicolumn{1}{c}{\shortstack{w/o \\ align}} & \multicolumn{1}{c}{\shortstack{w/ diversified \\ finetuning}} & \multicolumn{1}{c}{\shortstack{w/ naive \\ finetuning}} & \multicolumn{1}{c}{\shortstack{w/o \\ finetuning}} \\ 
        \midrule
        
        AR($\uparrow$)   & 55.7 & 54.2 & 32.9 & 0.0 & 52.4 & 11.4 & 12.6  \\ 
        CD($\downarrow$) & 0.72 & 0.94 & 1.05 & 38.47 & -- & -- & --  \\ 
        
    \bottomrule
    \end{tabular}
    \label{tab:ablation_table}
    }
\end{minipage}
\hspace{0.05\textwidth}
\begin{minipage}[t]{0.25\textwidth} 
\caption{Real-world evaluation.}
    \centering
    \begin{tabular}{lc}
        \toprule
        % Methods & SRT3D & DeepAC & {\cellcolor{gray!20} Ours} \\ 
        % Success Rate & 6.7 & 16.7 & {\rowcolor{gray!20}73.3} \\ 
        Methods & Success Rate($\uparrow$) \\
                \midrule
        SRT3D~\cite{stoiber2022srt3d} & 6.7 \\
        DeepAC~\cite{wang2023deep} &  16.7 \\
        \rowcolor{gray!20} Ours & 73.3 \\
    \bottomrule
    \end{tabular}
    \label{tab:in-the-wild}
\end{minipage}
\vspace{-0.6cm}
\end{table}

% \begin{table}[h]
% % \vspace{-0.6cm}
% \tiny
% \caption{ Real-world evaluation.}
%     \centering
%     \resizebox{0.3\columnwidth}{!}{
%     \begin{tabular}{lcccc}
%         \toprule
%         Methods & SRT3D & DeepAC & {\cellcolor{gray!20} Ours} \\ 
%         \midrule
%         Success Rate & 6.7 & 16.7 & {\rowcolor{gray!20}73.3} \\ 
%     \bottomrule
%     \end{tabular}
%     \label{tab:in-the-wild}

%     }
% \end{table}

% \begin{table}[t]
% \begin{minipage}[t]{0.4\textwidth} 
% \caption{ Effectiveness of training strategy.}
%     \centering
%     \begin{tabular}{lcccc}
%         \toprule
%         Methods & AR & MSSD & MSPD & VSD  \\ 
%         \midrule
%         Baseline & 12.6 & 20.8 & 9.6 & 7.3  \\ 
%         \textit{w/ naive finetuning} & 11.4 & 18.2 & 9.2 & 6.9  \\ 
%         {\rowcolor{gray!20} \textit{w/ new finetuning} & 52.4 & 49.5 & 51.5 & 56.1 }\\ 
%     \bottomrule
%     \end{tabular}
% \end{minipage}
% \hspace{0.05\textwidth}
% \begin{minipage}[t]{0.4\textwidth} 
% \caption{ Ablation Study.}
%     \centering
%     \begin{tabular}{lcccc}
%         \toprule
%         \multirow{2}{*}{Method} & \multicolumn{2}{c}{Metrix}\\ 
%          & AR($\uparrow$) & CD($\downarrow$)\\
%          \midrule
%          RosePose & 55.7 & 0.72 \\ 
%          \textit{w/o coarse align} & 54.2 & 0.94\\ 
%          \textit{w/o fine align} & 32.9 & 1.05\\ 
%         \textit{w/o align} & 0.0 & 38.47 \\
%     \bottomrule
%     \end{tabular}
% \end{minipage}

% \end{table}

\subsection{Ablation Study}
\label{ablation}
\textbf{Module Criticality Analysis.} We conducted ablation studies on our object alignment method. As shown in Table~\ref{tab:ablation_table}, the full method achieves effective alignment. Removing the coarse alignment stage (\textit{w/o coarse align}) led to a 1.5-point drop in AR and a 0.22 increase in CD. In contrast, removing the fine alignment stage (\textit{w/o fine align}) caused a much larger degradation: AR decreased by 22.8 points and CD increased by 0.33. These results confirm that both stages contribute to performance, with fine alignment playing a particularly critical role in achieving high-quality results.

\textbf{Efficacy of Synthetic Data Fine-Tuning.}  As shown in Table~\ref{tab:ablation_table}, we evaluate the impact of our text-guided generative augmentation on model fine-tuning on an annotated real-world dataset(Appendix.~\ref{annotated_dataset}). “Native fine-tuning” uses only the initial generated model, while “diversified fine-tuning” leverages texture-variant models from our generative strategy (Sec.~\ref{sec:finetune}). The latter yields a dramatic improvement: Average Recall (AR) jumps from 12.6\% (no fine-tuning) to 52.4\%—far outperforming native fine-tuning (11.4\%). This confirms that our synthetic data, by diversifying textures while preserving structural consistency, effectively bridges the domain gap between generated models and real-world images. Critically, this strategy requires only one reference image per novel object, eliminating the need for costly labeled datasets—making it uniquely practical for real-world deployment. More details can be seen in Appendix.~\ref{fitune_details}.

%===============================================================================

\section{Conclusion}
\label{sec:conclusion}
We introduced OnePoseViaGen, a novel approach for one-shot 6D pose estimation of unseen objects without requiring 3D models. Our method leverages single-view 3D generation and introduces a coarse-to-fine alignment strategy to simultaneously recover the object's metric scale and 6D pose using multi-view feature matching and refinement iteratively. To enhance robustness, we employ text-guided generative domain randomization to create diverse synthetic training data, effectively bridging the gap between generated models and real objects. Extensive evaluations on challenging benchmarks (YCBINEOAT, TOYL, LM-O) demonstrate state-of-the-art performance, significantly outperforming prior methods, especially under occlusion and varying illumination. Successful real-world robotic grasping experiments further validate the practical applicability of our system. OnePoseViaGen represents a significant step towards versatile, data-efficient pose estimation.

% ===============================================================================

\clearpage

\subsection{Limitations}
Although OnePoseViaGen achieves promising results across various datasets and real-world robotic experiments, it still faces challenges in handling deformable or articulated objects. In such cases, changes in object shape can lead to inaccurate 6D pose estimation. Future work will focus on incorporating test-time training into the inference pipeline, enabling continuous refinement of object geometry and accurate pose estimation for deformable objects. This direction aims to fully leverage the flexibility and generalization power of generative models in 6D pose estimation tasks.
% The acknowledgments are automatically included only in the final and preprint versions of the paper.
%===============================================================================

% no \bibliographystyle is required, since the corl style is automatically used.
\bibliography{main}  % .bib

\begin{thebibliography}{123}
\providecommand{\natexlab}[1]{#1}
\providecommand{\url}[1]{\texttt{#1}}
\expandafter\ifx\csname urlstyle\endcsname\relax
  \providecommand{\doi}[1]{doi: #1}\else
  \providecommand{\doi}{doi: \begingroup \urlstyle{rm}\Url}\fi

\bibitem[Chen et~al.(2024)Chen, Walsman, Memmel, Mo, Fang, Vemuri, Wu, Fox, and Gupta]{chen2024urdformer}
Z.~Chen, A.~Walsman, M.~Memmel, K.~Mo, A.~Fang, K.~Vemuri, A.~Wu, D.~Fox, and A.~Gupta.
\newblock Urdformer: A pipeline for constructing articulated simulation environments from real-world images.
\newblock \emph{arXiv preprint arXiv:2405.11656}, 2024.

\bibitem[Srinivasan et~al.(2024)Srinivasan, Collins, Heiden, Ng, Bohg, and Garg]{srinivasan2024dexmots}
K.~Srinivasan, J.~Collins, E.~Heiden, I.~Ng, J.~Bohg, and A.~Garg.
\newblock Dexmots: Dexterous manipulation with differentiable simulation.
\newblock 2024.

\bibitem[Dwivedi et~al.(2025)Dwivedi, Anti{\'c}, Tripathi, Taheri, Schmid, Black, and Tzionas]{dwivedi2025interactvlm}
S.~K. Dwivedi, D.~Anti{\'c}, S.~Tripathi, O.~Taheri, C.~Schmid, M.~J. Black, and D.~Tzionas.
\newblock Interactvlm: 3d interaction reasoning from 2d foundational models.
\newblock In \emph{Proceedings of the Computer Vision and Pattern Recognition Conference}, pages 22605--22615, 2025.

\bibitem[Zhong et~al.(2020)Zhong, Zhang, Zhao, Chang, Xiang, Zhang, and Zhang]{zhong2020seeing}
L.~Zhong, Y.~Zhang, H.~Zhao, A.~Chang, W.~Xiang, S.~Zhang, and L.~Zhang.
\newblock Seeing through the occluders: Robust monocular 6-dof object pose tracking via model-guided video object segmentation.
\newblock \emph{IEEE Robotics and Automation Letters}, 5\penalty0 (4):\penalty0 5159--5166, 2020.

\bibitem[Xia et~al.(2025)Xia, Su, Memmel, Jain, Yu, Mbiziwo-Tiapo, Farhadi, Gupta, Wang, and Ma]{xia2025drawer}
H.~Xia, E.~Su, M.~Memmel, A.~Jain, R.~Yu, N.~Mbiziwo-Tiapo, A.~Farhadi, A.~Gupta, S.~Wang, and W.-C. Ma.
\newblock Drawer: Digital reconstruction and articulation with environment realism.
\newblock In \emph{Proceedings of the Computer Vision and Pattern Recognition Conference}, pages 21771--21782, 2025.

\bibitem[Yang et~al.(2023)Yang, Ivanovic, Litany, Weng, Kim, Li, Che, Xu, Fidler, Pavone, et~al.]{yang2023emernerf}
J.~Yang, B.~Ivanovic, O.~Litany, X.~Weng, S.~W. Kim, B.~Li, T.~Che, D.~Xu, S.~Fidler, M.~Pavone, et~al.
\newblock Emernerf: Emergent spatial-temporal scene decomposition via self-supervision.
\newblock \emph{arXiv preprint arXiv:2311.02077}, 2023.

\bibitem[Xue et~al.(2023)Xue, Cheng, Kachana, and Xu]{xue2023neural}
S.~Xue, S.~Cheng, P.~Kachana, and D.~Xu.
\newblock Neural field dynamics model for granular object piles manipulation.
\newblock In \emph{Conference on Robot Learning}, pages 2821--2837. PMLR, 2023.

\bibitem[Chen et~al.(2025)Chen, Potamias, Chen, and Schmid]{chen2025hort}
Z.~Chen, R.~A. Potamias, S.~Chen, and C.~Schmid.
\newblock Hort: Monocular hand-held objects reconstruction with transformers.
\newblock \emph{arXiv preprint arXiv:2503.21313}, 2025.

\bibitem[Rashid et~al.(2023)Rashid, Sharma, Kim, Kerr, Chen, Kanazawa, and Goldberg]{rashid2023language}
A.~Rashid, S.~Sharma, C.~M. Kim, J.~Kerr, L.~Y. Chen, A.~Kanazawa, and K.~Goldberg.
\newblock Language embedded radiance fields for zero-shot task-oriented grasping.
\newblock In \emph{7th Annual Conference on Robot Learning}, 2023.

\bibitem[Wang et~al.(2024)Wang, Deng, Lum, Chen, Yang, Bohg, Zhu, and Guibas]{wang2024neural}
Q.~Wang, C.~Deng, T.~G.~W. Lum, Y.~Chen, Y.~Yang, J.~Bohg, Y.~Zhu, and L.~Guibas.
\newblock Neural attention field: Emerging point relevance in 3d scenes for one-shot dexterous grasping.
\newblock \emph{arXiv preprint arXiv:2410.23039}, 2024.

\bibitem[Wu et~al.(2016)Wu, Xue, Lim, Tian, Tenenbaum, Torralba, and Freeman]{wu2016single}
J.~Wu, T.~Xue, J.~J. Lim, Y.~Tian, J.~B. Tenenbaum, A.~Torralba, and W.~T. Freeman.
\newblock Single image 3d interpreter network.
\newblock In \emph{European Conference on Computer Vision}, pages 365--382. Springer, 2016.

\bibitem[Kulhanek et~al.(2024)Kulhanek, Peng, Kukelova, Pollefeys, and Sattler]{kulhanek2024wildgaussians}
J.~Kulhanek, S.~Peng, Z.~Kukelova, M.~Pollefeys, and T.~Sattler.
\newblock Wildgaussians: 3d gaussian splatting in the wild.
\newblock \emph{arXiv preprint arXiv:2407.08447}, 2024.

\bibitem[Matsuki et~al.(2024)Matsuki, Murai, Kelly, and Davison]{matsuki2024gaussian}
H.~Matsuki, R.~Murai, P.~H. Kelly, and A.~J. Davison.
\newblock Gaussian splatting slam.
\newblock In \emph{Proceedings of the IEEE/CVF Conference on Computer Vision and Pattern Recognition}, pages 18039--18048, 2024.

\bibitem[Dellaert et~al.(2020)Dellaert, Rosen, Wu, Mahony, and Carlone]{dellaert2020shonan}
F.~Dellaert, D.~M. Rosen, J.~Wu, R.~Mahony, and L.~Carlone.
\newblock Shonan rotation averaging: Global optimality by surfing so (p) n.
\newblock In \emph{European Conference on Computer Vision}, pages 292--308. Springer, 2020.

\bibitem[Zhou et~al.(2016)Zhou, Park, and Koltun]{zhou2016fast}
Q.-Y. Zhou, J.~Park, and V.~Koltun.
\newblock Fast global registration.
\newblock In \emph{European conference on computer vision}, pages 766--782. Springer, 2016.

\bibitem[Choy et~al.(2019)Choy, Park, and Koltun]{choy2019fully}
C.~Choy, J.~Park, and V.~Koltun.
\newblock Fully convolutional geometric features.
\newblock In \emph{Proceedings of the IEEE/CVF international conference on computer vision}, pages 8958--8966, 2019.

\bibitem[Simeonov et~al.(2023)Simeonov, Du, Lin, Garcia, Kaelbling, Lozano-P{\'e}rez, and Agrawal]{simeonov2023se}
A.~Simeonov, Y.~Du, Y.-C. Lin, A.~R. Garcia, L.~P. Kaelbling, T.~Lozano-P{\'e}rez, and P.~Agrawal.
\newblock Se (3)-equivariant relational rearrangement with neural descriptor fields.
\newblock In \emph{Conference on Robot Learning}, pages 835--846. PMLR, 2023.

\bibitem[Zeng et~al.(2017)Zeng, Song, Nie{\ss}ner, Fisher, Xiao, and Funkhouser]{zeng20173dmatch}
A.~Zeng, S.~Song, M.~Nie{\ss}ner, M.~Fisher, J.~Xiao, and T.~Funkhouser.
\newblock 3dmatch: Learning local geometric descriptors from rgb-d reconstructions.
\newblock In \emph{Proceedings of the IEEE conference on computer vision and pattern recognition}, pages 1802--1811, 2017.

\bibitem[Chen et~al.(2022)Chen, Zhao, Zhou, and Zhang]{chen2022pq}
X.~Chen, H.~Zhao, G.~Zhou, and Y.-Q. Zhang.
\newblock Pq-transformer: Jointly parsing 3d objects and layouts from point clouds.
\newblock \emph{IEEE Robotics and Automation Letters}, 7\penalty0 (2):\penalty0 2519--2526, 2022.

\bibitem[Wang et~al.(2025)Wang, Wang, Nakura, Bhowal, Kuo, Chen, Erickson, and Held]{wang2025articubot}
Y.~Wang, Z.~Wang, M.~Nakura, P.~Bhowal, C.-L. Kuo, Y.-T. Chen, Z.~Erickson, and D.~Held.
\newblock Articubot: Learning universal articulated object manipulation policy via large scale simulation.
\newblock \emph{arXiv preprint arXiv:2503.03045}, 2025.

\bibitem[Zeng et~al.(2017)Zeng, Yu, Song, Suo, Walker, Rodriguez, and Xiao]{zeng2017}
A.~Zeng, K.-T. Yu, S.~Song, D.~Suo, E.~Walker, A.~Rodriguez, and J.~Xiao.
\newblock Multi-view self-supervised deep learning for 6d pose estimation in the amazon picking challenge.
\newblock In \emph{2017 IEEE International Conference on Robotics and Automation (ICRA)}, pages 1386--1383, 2017.
\newblock \doi{10.1109/ICRA.2017.7989165}.

\bibitem[Kappler et~al.(2018)Kappler, Meier, Issac, Mainprice, Cifuentes, W{\"u}thrich, Berenz, Schaal, Ratliff, and Bohg]{kappler2018real}
D.~Kappler, F.~Meier, J.~Issac, J.~Mainprice, C.~G. Cifuentes, M.~W{\"u}thrich, V.~Berenz, S.~Schaal, N.~Ratliff, and J.~Bohg.
\newblock Real-time perception meets reactive motion generation.
\newblock \emph{IEEE Robotics and Automation Letters}, 3\penalty0 (3):\penalty0 1864--1871, 2018.

\bibitem[Wen et~al.(2022)Wen, Lian, Bekris, and Schaal]{wen2022catgrasp}
B.~Wen, W.~Lian, K.~Bekris, and S.~Schaal.
\newblock {CatGrasp}: Learning category-level task-relevant grasping in clutter from simulation.
\newblock In \emph{International Conference on Robotics and Automation (ICRA)}, pages 6401--6408, 2022.

\bibitem[Blukis et~al.(2023)Blukis, Lee, Tremblay, Wen, Kweon, Yoon, Fox, and Birchfield]{blukis2023oneshot}
V.~Blukis, T.~Lee, J.~Tremblay, B.~Wen, I.~S. Kweon, K.-J. Yoon, D.~Fox, and S.~Birchfield.
\newblock One-shot neural fields for {3D} object understanding.
\newblock 2023.

\bibitem[Zhuang et~al.(2023)Zhuang, Li, and Ding]{ZHUANG2023robobinpick}
C.~Zhuang, S.~Li, and H.~Ding.
\newblock Instance segmentation based 6d pose estimation of industrial objects using point clouds for robotic bin-picking.
\newblock \emph{Robotics and Computer-Integrated Manufacturing}, 82:\penalty0 102541, 2023.

\bibitem[Eisner et~al.(2024)Eisner, Yang, Davchev, Vecerik, Scholz, and Held]{eisner2024deep}
B.~Eisner, Y.~Yang, T.~Davchev, M.~Vecerik, J.~Scholz, and D.~Held.
\newblock Deep se (3)-equivariant geometric reasoning for precise placement tasks.
\newblock \emph{arXiv preprint arXiv:2404.13478}, 2024.

\bibitem[Zhao et~al.(2025)Zhao, Bogdanovic, Luo, Tohme, Darvish, Aspuru-Guzik, Shkurti, and Garg]{zhao2025anyplace}
Y.~Zhao, M.~Bogdanovic, C.~Luo, S.~Tohme, K.~Darvish, A.~Aspuru-Guzik, F.~Shkurti, and A.~Garg.
\newblock Anyplace: Learning generalized object placement for robot manipulation.
\newblock \emph{arXiv preprint arXiv:2502.04531}, 2025.

\bibitem[Mahler et~al.(2017)Mahler, Liang, Niyaz, Laskey, Doan, Liu, Ojea, and Goldberg]{mahler2017dex}
J.~Mahler, J.~Liang, S.~Niyaz, M.~Laskey, R.~Doan, X.~Liu, J.~A. Ojea, and K.~Goldberg.
\newblock Dex-net 2.0: Deep learning to plan robust grasps with synthetic point clouds and analytic grasp metrics.
\newblock \emph{arXiv preprint arXiv:1703.09312}, 2017.

\bibitem[Shen et~al.(2023)Shen, Yang, Yu, Wong, Kaelbling, and Isola]{shen2023distilled}
W.~Shen, G.~Yang, A.~Yu, J.~Wong, L.~P. Kaelbling, and P.~Isola.
\newblock Distilled feature fields enable few-shot language-guided manipulation.
\newblock \emph{arXiv preprint arXiv:2308.07931}, 2023.

\bibitem[Kerr et~al.(2024)Kerr, Kim, Wu, Yi, Wang, Goldberg, and Kanazawa]{kerr2024robot}
J.~Kerr, C.~M. Kim, M.~Wu, B.~Yi, Q.~Wang, K.~Goldberg, and A.~Kanazawa.
\newblock Robot see robot do: Imitating articulated object manipulation with monocular 4d reconstruction.
\newblock \emph{arXiv preprint arXiv:2409.18121}, 2024.

\bibitem[Caggiano et~al.()Caggiano, Dasari, and Kumar]{caggianomyodex}
V.~Caggiano, S.~Dasari, and V.~Kumar.
\newblock Myodex: Generalizable representations for dexterous physiological manipulation.

\bibitem[Chen et~al.(2022)Chen, Xu, and Agrawal]{chen2022system}
T.~Chen, J.~Xu, and P.~Agrawal.
\newblock A system for general in-hand object re-orientation.
\newblock In \emph{Conference on Robot Learning}, pages 297--307. PMLR, 2022.

\bibitem[Romero et~al.(2024)Romero, Fang, Agrawal, and Adelson]{romero2024eyesight}
B.~Romero, H.-S. Fang, P.~Agrawal, and E.~Adelson.
\newblock Eyesight hand: Design of a fully-actuated dexterous robot hand with integrated vision-based tactile sensors and compliant actuation.
\newblock In \emph{2024 IEEE/RSJ International Conference on Intelligent Robots and Systems (IROS)}, pages 1853--1860. IEEE, 2024.

\bibitem[Chen et~al.(2023)Chen, Tippur, Wu, Kumar, Adelson, and Agrawal]{chen2023visual}
T.~Chen, M.~Tippur, S.~Wu, V.~Kumar, E.~Adelson, and P.~Agrawal.
\newblock Visual dexterity: In-hand reorientation of novel and complex object shapes.
\newblock \emph{Science Robotics}, 8\penalty0 (84):\penalty0 eadc9244, 2023.

\bibitem[Bohg et~al.(2013)Bohg, Morales, Asfour, and Kragic]{bohg2013data}
J.~Bohg, A.~Morales, T.~Asfour, and D.~Kragic.
\newblock Data-driven grasp synthesis—a survey.
\newblock \emph{IEEE Transactions on robotics}, 30\penalty0 (2):\penalty0 289--309, 2013.

\bibitem[Driess et~al.(2023)Driess, Xia, Sajjadi, Lynch, Chowdhery, Wahid, Tompson, Vuong, Yu, Huang, et~al.]{driess2023palm}
D.~Driess, F.~Xia, M.~S. Sajjadi, C.~Lynch, A.~Chowdhery, A.~Wahid, J.~Tompson, Q.~Vuong, T.~Yu, W.~Huang, et~al.
\newblock Palm-e: An embodied multimodal language model.
\newblock 2023.

\bibitem[James et~al.(2020)James, Ma, Arrojo, and Davison]{james2020rlbench}
S.~James, Z.~Ma, D.~R. Arrojo, and A.~J. Davison.
\newblock Rlbench: The robot learning benchmark \& learning environment.
\newblock \emph{IEEE Robotics and Automation Letters}, 5\penalty0 (2):\penalty0 3019--3026, 2020.

\bibitem[Wei et~al.(2024)Wei, Xu, Guo, Hou, Gao, Cai, Luo, and Shao]{wei2024d}
Z.~Wei, Z.~Xu, J.~Guo, Y.~Hou, C.~Gao, Z.~Cai, J.~Luo, and L.~Shao.
\newblock D (r, o) grasp: A unified representation of robot and object interaction for cross-embodiment dexterous grasping.
\newblock \emph{arXiv preprint arXiv:2410.01702}, 2024.

\bibitem[Murali et~al.(2025)Murali, Sundaralingam, Chao, Yuan, Yamada, Carlson, Ramos, Birchfield, Fox, and Eppner]{murali2025graspgen}
A.~Murali, B.~Sundaralingam, Y.-W. Chao, W.~Yuan, J.~Yamada, M.~Carlson, F.~Ramos, S.~Birchfield, D.~Fox, and C.~Eppner.
\newblock Graspgen: A diffusion-based framework for 6-dof grasping with on-generator training.
\newblock \emph{arXiv preprint arXiv:2507.13097}, 2025.

\bibitem[Florence et~al.(2018)Florence, Manuelli, and Tedrake]{florence2018dense}
P.~R. Florence, L.~Manuelli, and R.~Tedrake.
\newblock Dense object nets: Learning dense visual object descriptors by and for robotic manipulation.
\newblock \emph{arXiv preprint arXiv:1806.08756}, 2018.

\bibitem[Billard and Kragic(2019)]{billard2019trends}
A.~Billard and D.~Kragic.
\newblock Trends and challenges in robot manipulation.
\newblock \emph{Science}, 364\penalty0 (6446):\penalty0 eaat8414, 2019.

\bibitem[Guo et~al.(2017)Guo, Sun, Liu, Kong, Fang, and Xi]{guo2017hybrid}
D.~Guo, F.~Sun, H.~Liu, T.~Kong, B.~Fang, and N.~Xi.
\newblock A hybrid deep architecture for robotic grasp detection.
\newblock In \emph{2017 IEEE international conference on robotics and automation (ICRA)}, pages 1609--1614. IEEE, 2017.

\bibitem[Deng et~al.(2019)Deng, Guo, Wei, Lu, Fang, Guo, Liu, and Sun]{deng2019deep}
Y.~Deng, X.~Guo, Y.~Wei, K.~Lu, B.~Fang, D.~Guo, H.~Liu, and F.~Sun.
\newblock Deep reinforcement learning for robotic pushing and picking in cluttered environment.
\newblock In \emph{2019 IEEE/RSJ International Conference on Intelligent Robots and Systems (IROS)}, pages 619--626. Ieee, 2019.

\bibitem[Sajjan et~al.(2020)Sajjan, Moore, Pan, Nagaraja, Lee, Zeng, and Song]{sajjan2020clear}
S.~Sajjan, M.~Moore, M.~Pan, G.~Nagaraja, J.~Lee, A.~Zeng, and S.~Song.
\newblock Clear grasp: 3d shape estimation of transparent objects for manipulation.
\newblock In \emph{2020 IEEE international conference on robotics and automation (ICRA)}, pages 3634--3642. IEEE, 2020.

\bibitem[Ding et~al.(2024)Ding, Chen, Wu, Li, Zhang, Gao, Li, Zhou, Zhu, Dong, et~al.]{ding2024preafford}
K.~Ding, B.~Chen, R.~Wu, Y.~Li, Z.~Zhang, H.-a. Gao, S.~Li, G.~Zhou, Y.~Zhu, H.~Dong, et~al.
\newblock Preafford: Universal affordance-based pre-grasping for diverse objects and environments.
\newblock In \emph{2024 IEEE/RSJ International Conference on Intelligent Robots and Systems (IROS)}, pages 7278--7285. IEEE, 2024.

\bibitem[Gervet et~al.(2023)Gervet, Chintala, Batra, Malik, and Chaplot]{gervet2023navigating}
T.~Gervet, S.~Chintala, D.~Batra, J.~Malik, and D.~S. Chaplot.
\newblock Navigating to objects in the real world.
\newblock \emph{Science Robotics}, 8\penalty0 (79):\penalty0 eadf6991, 2023.

\bibitem[Forster et~al.(2014)Forster, Pizzoli, and Scaramuzza]{forster2014svo}
C.~Forster, M.~Pizzoli, and D.~Scaramuzza.
\newblock Svo: Fast semi-direct monocular visual odometry.
\newblock In \emph{2014 IEEE international conference on robotics and automation (ICRA)}, pages 15--22. IEEE, 2014.

\bibitem[Kaufmann et~al.(2018)Kaufmann, Loquercio, Ranftl, Dosovitskiy, Koltun, and Scaramuzza]{kaufmann2018deep}
E.~Kaufmann, A.~Loquercio, R.~Ranftl, A.~Dosovitskiy, V.~Koltun, and D.~Scaramuzza.
\newblock Deep drone racing: Learning agile flight in dynamic environments.
\newblock In \emph{Conference on Robot Learning}, pages 133--145. PMLR, 2018.

\bibitem[Sabbah et~al.(2025)Sabbah, Wojciechowski, Soh, Hsu, Righetti, Mansard, Watier, and Bonnet]{sabbah2025optimal}
M.~Sabbah, K.~Wojciechowski, H.~Soh, D.~Hsu, L.~Righetti, N.~Mansard, B.~Watier, and V.~Bonnet.
\newblock Optimal motion prediction for human-to-robot handovers.
\newblock 2025.

\bibitem[Runz et~al.(2018)Runz, Buffier, and Agapito]{runz2018maskfusion}
M.~Runz, M.~Buffier, and L.~Agapito.
\newblock {MaskFusion}: Real-time recognition, tracking and reconstruction of multiple moving objects.
\newblock 2018.

\bibitem[Marchand et~al.(2015)Marchand, Uchiyama, and Spindler]{marchand2015pose}
E.~Marchand, H.~Uchiyama, and F.~Spindler.
\newblock Pose estimation for augmented reality: A hands-on survey.
\newblock \emph{IEEE Transactions on Visualization and Computer Graphics (TVCG)}, 22\penalty0 (12):\penalty0 2633--2651, 2015.

\bibitem[Kim et~al.(2024)Kim, Pertsch, Karamcheti, Xiao, Balakrishna, Nair, Rafailov, Foster, Lam, Sanketi, et~al.]{kim2024openvla}
M.~J. Kim, K.~Pertsch, S.~Karamcheti, T.~Xiao, A.~Balakrishna, S.~Nair, R.~Rafailov, E.~Foster, G.~Lam, P.~Sanketi, et~al.
\newblock Openvla: An open-source vision-language-action model.
\newblock \emph{arXiv preprint arXiv:2406.09246}, 2024.

\bibitem[Team et~al.(2024)Team, Ghosh, Walke, Pertsch, Black, Mees, Dasari, Hejna, Kreiman, Xu, et~al.]{team2024octo}
O.~M. Team, D.~Ghosh, H.~Walke, K.~Pertsch, K.~Black, O.~Mees, S.~Dasari, J.~Hejna, T.~Kreiman, C.~Xu, et~al.
\newblock Octo: An open-source generalist robot policy.
\newblock \emph{arXiv preprint arXiv:2405.12213}, 2024.

\bibitem[Venkataraman et~al.(2024)Venkataraman, Wang, Wang, Erickson, and Held]{venkataraman2024real}
S.~Venkataraman, Y.~Wang, Z.~Wang, Z.~Erickson, and D.~Held.
\newblock Real-world offline reinforcement learning from vision language model feedback.
\newblock \emph{arXiv preprint arXiv:2411.05273}, 2024.

\bibitem[Ahn et~al.(2022)Ahn, Brohan, Brown, Chebotar, Cortes, David, Finn, Fu, Gopalakrishnan, Hausman, et~al.]{ahn2022can}
M.~Ahn, A.~Brohan, N.~Brown, Y.~Chebotar, O.~Cortes, B.~David, C.~Finn, C.~Fu, K.~Gopalakrishnan, K.~Hausman, et~al.
\newblock Do as i can, not as i say: Grounding language in robotic affordances.
\newblock \emph{arXiv preprint arXiv:2204.01691}, 2022.

\bibitem[Zitkovich et~al.(2023)Zitkovich, Yu, Xu, Xu, Xiao, Xia, Wu, Wohlhart, Welker, Wahid, et~al.]{zitkovich2023rt}
B.~Zitkovich, T.~Yu, S.~Xu, P.~Xu, T.~Xiao, F.~Xia, J.~Wu, P.~Wohlhart, S.~Welker, A.~Wahid, et~al.
\newblock Rt-2: Vision-language-action models transfer web knowledge to robotic control.
\newblock In \emph{Conference on Robot Learning}, pages 2165--2183. PMLR, 2023.

\bibitem[Ma et~al.(2023)Ma, Liang, Wang, Huang, Bastani, Jayaraman, Zhu, Fan, and Anandkumar]{ma2023eureka}
Y.~J. Ma, W.~Liang, G.~Wang, D.-A. Huang, O.~Bastani, D.~Jayaraman, Y.~Zhu, L.~Fan, and A.~Anandkumar.
\newblock Eureka: Human-level reward design via coding large language models.
\newblock \emph{arXiv preprint arXiv:2310.12931}, 2023.

\bibitem[O’Neill et~al.(2024)O’Neill, Rehman, Maddukuri, Gupta, Padalkar, Lee, Pooley, Gupta, Mandlekar, Jain, et~al.]{o2024open}
A.~O’Neill, A.~Rehman, A.~Maddukuri, A.~Gupta, A.~Padalkar, A.~Lee, A.~Pooley, A.~Gupta, A.~Mandlekar, A.~Jain, et~al.
\newblock Open x-embodiment: Robotic learning datasets and rt-x models: Open x-embodiment collaboration 0.
\newblock In \emph{2024 IEEE International Conference on Robotics and Automation (ICRA)}, pages 6892--6903. IEEE, 2024.

\bibitem[Khazatsky et~al.(2024)Khazatsky, Pertsch, Nair, Balakrishna, Dasari, Karamcheti, Nasiriany, Srirama, Chen, Ellis, et~al.]{khazatsky2024droid}
A.~Khazatsky, K.~Pertsch, S.~Nair, A.~Balakrishna, S.~Dasari, S.~Karamcheti, S.~Nasiriany, M.~K. Srirama, L.~Y. Chen, K.~Ellis, et~al.
\newblock Droid: A large-scale in-the-wild robot manipulation dataset.
\newblock \emph{arXiv preprint arXiv:2403.12945}, 2024.

\bibitem[Kumar et~al.(2023)Kumar, Shah, Zhou, Moens, Caggiano, Gupta, and Rajeswaran]{kumar2023robohive}
V.~Kumar, R.~Shah, G.~Zhou, V.~Moens, V.~Caggiano, A.~Gupta, and A.~Rajeswaran.
\newblock Robohive: A unified framework for robot learning.
\newblock \emph{Advances in Neural Information Processing Systems}, 36:\penalty0 44323--44340, 2023.

\bibitem[Chi et~al.(2023)Chi, Xu, Feng, Cousineau, Du, Burchfiel, Tedrake, and Song]{chi2023diffusion}
C.~Chi, Z.~Xu, S.~Feng, E.~Cousineau, Y.~Du, B.~Burchfiel, R.~Tedrake, and S.~Song.
\newblock Diffusion policy: Visuomotor policy learning via action diffusion.
\newblock \emph{The International Journal of Robotics Research}, page 02783649241273668, 2023.

\bibitem[Kehl et~al.(2017)Kehl, Manhardt, Tombari, Ilic, and Navab]{kehl2017ssd6d}
W.~Kehl, F.~Manhardt, F.~Tombari, S.~Ilic, and N.~Navab.
\newblock { SSD-6D: Making RGB-Based 3D Detection and 6D Pose Estimation Great Again }.
\newblock In \emph{2017 IEEE International Conference on Computer Vision (ICCV)}, pages 1530--1538, Los Alamitos, CA, USA, Oct. 2017. IEEE Computer Society.
\newblock \doi{10.1109/ICCV.2017.169}.
\newblock URL \url{https://doi.ieeecomputersociety.org/10.1109/ICCV.2017.169}.

\bibitem[Peng et~al.(2019)Peng, Liu, Huang, Zhou, and Bao]{peng2019pvnet}
S.~Peng, Y.~Liu, Q.~Huang, X.~Zhou, and H.~Bao.
\newblock Pvnet: Pixel-wise voting network for 6dof pose estimation.
\newblock In \emph{CVPR}, 2019.

\bibitem[Wen et~al.(2020)Wen, Mitash, Soorian, Kimmel, Sintov, and Bekris]{wen2020ropose}
B.~Wen, C.~Mitash, S.~Soorian, A.~Kimmel, A.~Sintov, and K.~E. Bekris.
\newblock Robust, occlusion-aware pose estimation for objects grasped by adaptive hands.
\newblock In \emph{2020 IEEE International Conference on Robotics and Automation (ICRA)}, pages 6210--6217, 2020.
\newblock \doi{10.1109/ICRA40945.2020.9197350}.

\bibitem[Lim et~al.(2013)Lim, Pirsiavash, and Torralba]{lim2013parsing}
J.~J. Lim, H.~Pirsiavash, and A.~Torralba.
\newblock Parsing ikea objects: Fine pose estimation.
\newblock In \emph{Proceedings of the IEEE international conference on computer vision}, pages 2992--2999, 2013.

\bibitem[Wang et~al.(2019)Wang, Xu, Zhu, Mart{\'\i}n-Mart{\'\i}n, Lu, Fei-Fei, and Savarese]{wang2019densefusion}
C.~Wang, D.~Xu, Y.~Zhu, R.~Mart{\'\i}n-Mart{\'\i}n, C.~Lu, L.~Fei-Fei, and S.~Savarese.
\newblock Densefusion: 6d object pose estimation by iterative dense fusion.
\newblock In \emph{Proceedings of the IEEE/CVF conference on computer vision and pattern recognition}, pages 3343--3352, 2019.

\bibitem[Chi and Song(2021)]{Chi_2021_ICCV}
C.~Chi and S.~Song.
\newblock Garmentnets: Category-level pose estimation for garments via canonical space shape completion.
\newblock In \emph{Proceedings of the IEEE/CVF International Conference on Computer Vision (ICCV)}, pages 3324--3333, October 2021.

\bibitem[Li et~al.(2020)Li, Wang, Yi, Guibas, Abbott, and Song]{Li_2020_CVPR}
X.~Li, H.~Wang, L.~Yi, L.~J. Guibas, A.~L. Abbott, and S.~Song.
\newblock Category-level articulated object pose estimation.
\newblock In \emph{Proceedings of the IEEE/CVF Conference on Computer Vision and Pattern Recognition (CVPR)}, June 2020.

\bibitem[Goodwin et~al.(2022)Goodwin, Vaze, Havoutis, and Posner]{goodwin2022a}
W.~Goodwin, S.~Vaze, I.~Havoutis, and I.~Posner.
\newblock Zero-shot category-level object pose estimation.
\newblock Number 13699 in Lecture Notes in Computer Science, pages 516--532. Springer, 2022.

\bibitem[Li et~al.(2023)Li, Zhu, Zhang, Shi, Zhang, Zhang, and Li]{li2023sd}
G.~Li, D.~Zhu, G.~Zhang, W.~Shi, T.~Zhang, X.~Zhang, and J.~Li.
\newblock Sd-pose: structural discrepancy aware category-level 6d object pose estimation.
\newblock In \emph{Proceedings of the IEEE/CVF Winter Conference on applications of computer vision}, pages 5685--5694, 2023.

\bibitem[Lin et~al.(2021)Lin, Wei, Li, Xu, Jia, and Li]{lin2021dualposenet}
J.~Lin, Z.~Wei, Z.~Li, S.~Xu, K.~Jia, and Y.~Li.
\newblock Dualposenet: Category-level 6d object pose and size estimation using dual pose network with refined learning of pose consistency.
\newblock In \emph{Proceedings of the IEEE/CVF International Conference on Computer Vision}, pages 3560--3569, 2021.

\bibitem[Nguyen et~al.(2024)Nguyen, Groueix, Salzmann, and Lepetit]{nguyen2024gigapose}
V.~N. Nguyen, T.~Groueix, M.~Salzmann, and V.~Lepetit.
\newblock Gigapose: Fast and robust novel object pose estimation via one correspondence.
\newblock In \emph{Proceedings of the IEEE/CVF Conference on Computer Vision and Pattern Recognition}, pages 9903--9913, 2024.

\bibitem[{\"O}rnek et~al.(2024){\"O}rnek, Labb{\'e}, Tekin, Ma, Keskin, Forster, and Hodan]{ornek2024foundpose}
E.~P. {\"O}rnek, Y.~Labb{\'e}, B.~Tekin, L.~Ma, C.~Keskin, C.~Forster, and T.~Hodan.
\newblock Foundpose: Unseen object pose estimation with foundation features.
\newblock In \emph{European Conference on Computer Vision}, pages 163--182. Springer, 2024.

\bibitem[Wen et~al.(2024)Wen, Yang, Kautz, and Birchfield]{wen2024foundationpose}
B.~Wen, W.~Yang, J.~Kautz, and S.~Birchfield.
\newblock Foundationpose: Unified 6d pose estimation and tracking of novel objects.
\newblock In \emph{Proceedings of the IEEE/CVF Conference on Computer Vision and Pattern Recognition}, pages 17868--17879, 2024.

\bibitem[He et~al.(2022)He, Wang, Fan, Sun, and Chen]{he2022fs6d}
Y.~He, Y.~Wang, H.~Fan, J.~Sun, and Q.~Chen.
\newblock Fs6d: Few-shot 6d pose estimation of novel objects.
\newblock In \emph{Proceedings of the IEEE/CVF Conference on Computer Vision and Pattern Recognition}, pages 6814--6824, 2022.

\bibitem[Labb{\'e} et~al.(2022)Labb{\'e}, Manuelli, Mousavian, Tyree, Birchfield, Tremblay, Carpentier, Aubry, Fox, and Sivic]{labbe2022megapose}
Y.~Labb{\'e}, L.~Manuelli, A.~Mousavian, S.~Tyree, S.~Birchfield, J.~Tremblay, J.~Carpentier, M.~Aubry, D.~Fox, and J.~Sivic.
\newblock Megapose: 6d pose estimation of novel objects via render \& compare.
\newblock \emph{arXiv preprint arXiv:2212.06870}, 2022.

\bibitem[Moon et~al.(2025)Moon, Son, Hur, and Kim]{moon2025co}
S.~Moon, H.~Son, D.~Hur, and S.~Kim.
\newblock Co-op: Correspondence-based novel object pose estimation.
\newblock In \emph{Proceedings of the IEEE/CVF Conference on Computer Vision and Pattern Recognition}, 2025.

\bibitem[Sun et~al.(2021)Sun, Shen, Wang, Bao, and Zhou]{sun2021loftr}
J.~Sun, Z.~Shen, Y.~Wang, H.~Bao, and X.~Zhou.
\newblock Loftr: Detector-free local feature matching with transformers.
\newblock In \emph{Proceedings of the IEEE/CVF conference on computer vision and pattern recognition}, pages 8922--8931, 2021.

\bibitem[Corsetti et~al.(2024)Corsetti, Boscaini, Oh, Cavallaro, and Poiesi]{corsetti2024open}
J.~Corsetti, D.~Boscaini, C.~Oh, A.~Cavallaro, and F.~Poiesi.
\newblock Open-vocabulary object 6d pose estimation.
\newblock In \emph{Proceedings of the IEEE/CVF Conference on Computer Vision and Pattern Recognition}, pages 18071--18080, 2024.

\bibitem[Liu et~al.(2024)Liu, Wang, Zhang, Zhang, Tombari, and Ji]{liu2024unopose}
X.~Liu, G.~Wang, R.~Zhang, C.~Zhang, F.~Tombari, and X.~Ji.
\newblock Unopose: Unseen object pose estimation with an unposed rgb-d reference image.
\newblock \emph{arXiv preprint arXiv:2411.16106}, 2024.

\bibitem[Jin et~al.(2024)Jin, Prasad, Jauhri, Franzius, and Chalvatzaki]{jin20246dope}
Y.~Jin, V.~Prasad, S.~Jauhri, M.~Franzius, and G.~Chalvatzaki.
\newblock 6dope-gs: Online 6d object pose estimation using gaussian splatting.
\newblock \emph{arXiv preprint arXiv:2412.01543}, 2024.

\bibitem[Ye et~al.(2025)Ye, Wu, Lu, Chang, Guo, Zhou, Zhao, and Han]{ye2025hi3dgen}
C.~Ye, Y.~Wu, Z.~Lu, J.~Chang, X.~Guo, J.~Zhou, H.~Zhao, and X.~Han.
\newblock Hi3dgen: High-fidelity 3d geometry generation from images via normal bridging.
\newblock \emph{arXiv preprint arXiv:2503.22236}, 3, 2025.

\bibitem[Guo et~al.(2025)Guo, Tang, Cha, Zhang, Liu, and Wu]{guo2025craft}
M.~Guo, M.~Tang, H.~Cha, R.~Zhang, C.~K. Liu, and J.~Wu.
\newblock Craft: Designing creative and functional 3d objects.
\newblock In \emph{2025 IEEE/CVF Winter Conference on Applications of Computer Vision (WACV)}, pages 7215--7224. IEEE, 2025.

\bibitem[Xiang et~al.(2024)Xiang, Lv, Xu, Deng, Wang, Zhang, Chen, Tong, and Yang]{xiang2024trellis}
J.~Xiang, Z.~Lv, S.~Xu, Y.~Deng, R.~Wang, B.~Zhang, D.~Chen, X.~Tong, and J.~Yang.
\newblock Structured 3d latents for scalable and versatile 3d generation.
\newblock \emph{arXiv preprint arXiv:2412.01506}, 2024.

\bibitem[Lim et~al.(2013)Lim, Pirsiavash, and Torralba]{Lim_2013_ICCV}
J.~J. Lim, H.~Pirsiavash, and A.~Torralba.
\newblock Parsing ikea objects: Fine pose estimation.
\newblock In \emph{Proceedings of the IEEE International Conference on Computer Vision (ICCV)}, December 2013.

\bibitem[Wang et~al.(2019)Wang, Sridhar, Huang, Valentin, Song, and Guibas]{Wang_2019_CVPR}
H.~Wang, S.~Sridhar, J.~Huang, J.~Valentin, S.~Song, and L.~J. Guibas.
\newblock Normalized object coordinate space for category-level 6d object pose and size estimation.
\newblock In \emph{Proceedings of the IEEE/CVF Conference on Computer Vision and Pattern Recognition (CVPR)}, June 2019.

\bibitem[Liu et~al.(2022)Liu, Wen, Peng, Lin, Long, Komura, and Wang]{liu2022gen6d}
Y.~Liu, Y.~Wen, S.~Peng, C.~Lin, X.~Long, T.~Komura, and W.~Wang.
\newblock Gen6d: Generalizable model-free 6-dof object pose estimation from rgb images.
\newblock In \emph{European Conference on Computer Vision}, pages 298--315. Springer, 2022.

\bibitem[Sun et~al.(2022)Sun, Wang, Zhang, He, Zhao, Zhang, and Zhou]{sun2022onepose}
J.~Sun, Z.~Wang, S.~Zhang, X.~He, H.~Zhao, G.~Zhang, and X.~Zhou.
\newblock Onepose: One-shot object pose estimation without cad models.
\newblock In \emph{Proceedings of the IEEE/CVF Conference on Computer Vision and Pattern Recognition}, pages 6825--6834, 2022.

\bibitem[Huang et~al.(2023)Huang, Ahn, Li, Hu, and Lee]{Huang2023}
D.~Huang, H.~Ahn, S.~Li, Y.~Hu, and D.~Lee.
\newblock Estimation of 6d pose of objects based on a variant adversarial autoencoder.
\newblock \emph{Neural Processing Letters}, 55\penalty0 (7):\penalty0 9581--9596, Dec 2023.
\newblock ISSN 1573-773X.
\newblock \doi{10.1007/s11063-023-11215-2}.
\newblock URL \url{https://doi.org/10.1007/s11063-023-11215-2}.

\bibitem[Zhong et~al.(2023)Zhong, Zheng, Zheng, Zhao, Yi, Mu, Wang, Li, Zhou, Yang, et~al.]{zhong20233d}
C.~Zhong, Y.~Zheng, Y.~Zheng, H.~Zhao, L.~Yi, X.~Mu, L.~Wang, P.~Li, G.~Zhou, C.~Yang, et~al.
\newblock 3d implicit transporter for temporally consistent keypoint discovery.
\newblock In \emph{Proceedings of the IEEE/CVF international conference on computer vision}, pages 3869--3880, 2023.

\bibitem[Zhong et~al.(2022)Zhong, You, Chen, Zhao, Sun, Zhou, Mu, Gan, and Huang]{zhong2022snake}
C.~Zhong, P.~You, X.~Chen, H.~Zhao, F.~Sun, G.~Zhou, X.~Mu, C.~Gan, and W.~Huang.
\newblock Snake: Shape-aware neural 3d keypoint field.
\newblock \emph{Advances in Neural Information Processing Systems}, 35:\penalty0 7052--7064, 2022.

\bibitem[Oquab et~al.(2023)Oquab, Darcet, Moutakanni, Vo, Szafraniec, Khalidov, Fernandez, Haziza, Massa, El-Nouby, et~al.]{oquab2023dinov2}
M.~Oquab, T.~Darcet, T.~Moutakanni, H.~Vo, M.~Szafraniec, V.~Khalidov, P.~Fernandez, D.~Haziza, F.~Massa, A.~El-Nouby, et~al.
\newblock Dinov2: Learning robust visual features without supervision.
\newblock \emph{arXiv preprint arXiv:2304.07193}, 2023.

\bibitem[He et~al.(2022)He, Sun, Wang, Huang, Bao, and Zhou]{he2022onepose++}
X.~He, J.~Sun, Y.~Wang, D.~Huang, H.~Bao, and X.~Zhou.
\newblock Onepose++: Keypoint-free one-shot object pose estimation without cad models.
\newblock \emph{Advances in Neural Information Processing Systems}, 35:\penalty0 35103--35115, 2022.

\bibitem[Stoiber et~al.(2022)Stoiber, Pfanne, Strobl, Triebel, and Albu-Sch{\"a}ffer]{stoiber2022srt3d}
M.~Stoiber, M.~Pfanne, K.~H. Strobl, R.~Triebel, and A.~Albu-Sch{\"a}ffer.
\newblock Srt3d: A sparse region-based 3d object tracking approach for the real world.
\newblock \emph{International Journal of Computer Vision}, 130\penalty0 (4):\penalty0 1008--1030, 2022.

\bibitem[Xiang et~al.(2024)Xiang, Lv, Xu, Deng, Wang, Zhang, Chen, Tong, and Yang]{xiang2024structured}
J.~Xiang, Z.~Lv, S.~Xu, Y.~Deng, R.~Wang, B.~Zhang, D.~Chen, X.~Tong, and J.~Yang.
\newblock Structured 3d latents for scalable and versatile 3d generation.
\newblock \emph{arXiv preprint arXiv:2412.01506}, 2024.

\bibitem[Chen et~al.(2024{\natexlab{a}})Chen, Ding, Zhang, Yu, Zang, Li, Peng, and Sun]{chen2024rapid}
T.~Chen, C.~Ding, S.~Zhang, C.~Yu, Y.~Zang, Z.~Li, S.~Peng, and L.~Sun.
\newblock Rapid 3d model generation with intuitive 3d input.
\newblock In \emph{Proceedings of the IEEE/CVF Conference on Computer Vision and Pattern Recognition}, pages 12554--12564, 2024{\natexlab{a}}.

\bibitem[Chen et~al.(2024{\natexlab{b}})Chen, Pan, Yang, Yao, and Mei]{chen2024vp3d}
Y.~Chen, Y.~Pan, H.~Yang, T.~Yao, and T.~Mei.
\newblock Vp3d: Unleashing 2d visual prompt for text-to-3d generation.
\newblock In \emph{Proceedings of the IEEE/CVF Conference on Computer Vision and Pattern Recognition}, pages 4896--4905, 2024{\natexlab{b}}.

\bibitem[Chen et~al.(2024{\natexlab{c}})Chen, Zhang, Yang, Cai, Yu, Yang, and Lin]{chen2024it3d}
Y.~Chen, C.~Zhang, X.~Yang, Z.~Cai, G.~Yu, L.~Yang, and G.~Lin.
\newblock It3d: Improved text-to-3d generation with explicit view synthesis.
\newblock In \emph{Proceedings of the AAAI Conference on Artificial Intelligence}, volume~38, pages 1237--1244, 2024{\natexlab{c}}.

\bibitem[Tang et~al.(2025)Tang, Zhang, Cheng, Yu, Feng, Pang, Lin, and Yuan]{tang2025cycle3d}
Z.~Tang, J.~Zhang, X.~Cheng, W.~Yu, C.~Feng, Y.~Pang, B.~Lin, and L.~Yuan.
\newblock Cycle3d: High-quality and consistent image-to-3d generation via generation-reconstruction cycle.
\newblock In \emph{Proceedings of the AAAI Conference on Artificial Intelligence}, volume~39, pages 7320--7328, 2025.

\bibitem[Xie et~al.(2024)Xie, Zhang, Tang, Wu, Chen, Li, and Jin]{xie2024styletex}
Z.~Xie, Y.~Zhang, X.~Tang, Y.~Wu, D.~Chen, G.~Li, and X.~Jin.
\newblock Styletex: Style image-guided texture generation for 3d models.
\newblock \emph{ACM Transactions on Graphics (TOG)}, 43\penalty0 (6):\penalty0 1--14, 2024.

\bibitem[Wu et~al.(2024)Wu, Zhou, Yi, Yuan, and Zhang]{wu2024consistent3d}
Z.~Wu, P.~Zhou, X.~Yi, X.~Yuan, and H.~Zhang.
\newblock Consistent3d: Towards consistent high-fidelity text-to-3d generation with deterministic sampling prior.
\newblock In \emph{Proceedings of the IEEE/CVF Conference on Computer Vision and Pattern Recognition}, pages 9892--9902, 2024.

\bibitem[Xu et~al.(2024)Xu, Chen, Tu, Gong, Liu, Mei, Ren, and Li]{xu20243d}
R.~Xu, R.~Chen, C.~Tu, X.~Gong, Z.~Liu, L.~Mei, X.~Ren, and Z.~Li.
\newblock 3d models of sarcomas: the next-generation tool for personalized medicine.
\newblock \emph{Phenomics}, 4\penalty0 (2):\penalty0 171--186, 2024.

\bibitem[M\"uller et~al.(2022)M\"uller, Evans, Schied, and Keller]{mueller2022instant}
T.~M\"uller, A.~Evans, C.~Schied, and A.~Keller.
\newblock Instant neural graphics primitives with a multiresolution hash encoding.
\newblock \emph{ACM Trans. Graph.}, 41\penalty0 (4):\penalty0 102:1--102:15, July 2022.
\newblock \doi{10.1145/3528223.3530127}.
\newblock URL \url{https://doi.org/10.1145/3528223.3530127}.

\bibitem[Wang et~al.(2021)Wang, Liu, Liu, Theobalt, Komura, and Wang]{wang2021neus}
P.~Wang, L.~Liu, Y.~Liu, C.~Theobalt, T.~Komura, and W.~Wang.
\newblock {NeuS}: Learning neural implicit surfaces by volume rendering for multi-view reconstruction.
\newblock In \emph{Advances in Neural Information Processing Systems (NeurIPS)}, 2021.

\bibitem[Wen et~al.(2023)Wen, Tremblay, Blukis, Tyree, M{\"u}ller, Evans, Fox, Kautz, and Birchfield]{wen2023bundlesdf}
B.~Wen, J.~Tremblay, V.~Blukis, S.~Tyree, T.~M{\"u}ller, A.~Evans, D.~Fox, J.~Kautz, and S.~Birchfield.
\newblock Bundlesdf: Neural 6-dof tracking and 3d reconstruction of unknown objects.
\newblock In \emph{Proceedings of the IEEE/CVF Conference on Computer Vision and Pattern Recognition}, pages 606--617, 2023.

\bibitem[DeTone et~al.(2018)DeTone, Malisiewicz, and Rabinovich]{detone2018superpoint}
D.~DeTone, T.~Malisiewicz, and A.~Rabinovich.
\newblock { SuperPoint: Self-Supervised Interest Point Detection and Description }.
\newblock In \emph{2018 IEEE/CVF Conference on Computer Vision and Pattern Recognition Workshops (CVPRW)}, pages 337--33712, Los Alamitos, CA, USA, June 2018. IEEE Computer Society.
\newblock \doi{10.1109/CVPRW.2018.00060}.
\newblock URL \url{https://doi.ieeecomputersociety.org/10.1109/CVPRW.2018.00060}.

\bibitem[Sarlin et~al.(2020)Sarlin, DeTone, Malisiewicz, and Rabinovich]{Sarlin2020superglue}
P.-E. Sarlin, D.~DeTone, T.~Malisiewicz, and A.~Rabinovich.
\newblock Superglue: Learning feature matching with graph neural networks.
\newblock In \emph{2020 IEEE/CVF Conference on Computer Vision and Pattern Recognition (CVPR)}, pages 4937--4946, 2020.

\bibitem[Fischler and Bolles(1987)]{MartinPnP1987}
M.~A. Fischler and R.~C. Bolles.
\newblock Random sample consensus: A paradigm for model fitting with applications to image analysis and automated cartography.
\newblock In \emph{Readings in Computer Vision}, pages 726--740. Morgan Kaufmann, 1987.

\bibitem[Poiesi and Boscaini(2022)]{poiesi2022learning}
F.~Poiesi and D.~Boscaini.
\newblock Learning general and distinctive 3d local deep descriptors for point cloud registration.
\newblock \emph{IEEE Transactions on Pattern Analysis and Machine Intelligence}, 45\penalty0 (3):\penalty0 3979--3985, 2022.

\bibitem[Lee et~al.(2025)Lee, Wen, Kang, Kang, Kweon, and Yoon]{lee2025any6d}
T.~Lee, B.~Wen, M.~Kang, G.~Kang, I.~S. Kweon, and K.-J. Yoon.
\newblock {Any6D}: Model-free 6d pose estimation of novel objects.
\newblock In \emph{Proceedings of the Computer Vision and Pattern Recognition Conference (CVPR)}, 2025.

\bibitem[Lowe(1999)]{lowe1999object}
D.~G. Lowe.
\newblock Object recognition from local scale-invariant features.
\newblock 1999.

\bibitem[G{\"u}meli et~al.(2023)G{\"u}meli, Dai, and Nie{\ss}ner]{gumeli2023objectmatch}
C.~G{\"u}meli, A.~Dai, and M.~Nie{\ss}ner.
\newblock Objectmatch: Robust registration using canonical object correspondences.
\newblock 2023.

\bibitem[Wen et~al.(2021)Wen, Mitash, and Bekris]{wen2021data}
B.~Wen, C.~Mitash, and K.~Bekris.
\newblock Data-driven 6d pose tracking by calibrating image residuals in synthetic domains.
\newblock \emph{arXiv preprint arXiv:2105.14391}, 2021.

\bibitem[Hodan et~al.(2018)Hodan, Michel, Brachmann, Kehl, GlentBuch, Kraft, Drost, Vidal, Ihrke, Zabulis, et~al.]{hodan2018bop}
T.~Hodan, F.~Michel, E.~Brachmann, W.~Kehl, A.~GlentBuch, D.~Kraft, B.~Drost, J.~Vidal, S.~Ihrke, X.~Zabulis, et~al.
\newblock Bop: Benchmark for 6d object pose estimation.
\newblock In \emph{Proceedings of the European conference on computer vision (ECCV)}, pages 19--34, 2018.

\bibitem[Brachmann et~al.(2014)Brachmann, Krull, Michel, Gumhold, Shotton, and Rother]{brachmann2014learning}
E.~Brachmann, A.~Krull, F.~Michel, S.~Gumhold, J.~Shotton, and C.~Rother.
\newblock Learning 6d object pose estimation using 3d object coordinates.
\newblock In \emph{Computer Vision--ECCV 2014: 13th European Conference, Zurich, Switzerland, September 6-12, 2014, Proceedings, Part II 13}, pages 536--551. Springer, 2014.

\bibitem[Chao et~al.(2021)Chao, Yang, Xiang, Molchanov, Handa, Tremblay, Narang, Van~Wyk, Iqbal, Birchfield, et~al.]{chao2021dexycb}
Y.-W. Chao, W.~Yang, Y.~Xiang, P.~Molchanov, A.~Handa, J.~Tremblay, Y.~S. Narang, K.~Van~Wyk, U.~Iqbal, S.~Birchfield, et~al.
\newblock Dexycb: A benchmark for capturing hand grasping of objects.
\newblock In \emph{Proceedings of the IEEE/CVF conference on computer vision and pattern recognition}, pages 9044--9053, 2021.

\bibitem[Wang et~al.(2023)Wang, Yan, Zhen, Liu, Zhang, Zhang, and Zhou]{wang2023deep}
L.~Wang, S.~Yan, J.~Zhen, Y.~Liu, M.~Zhang, G.~Zhang, and X.~Zhou.
\newblock Deep active contours for real-time 6-dof object tracking.
\newblock In \emph{Proceedings of the IEEE/CVF International Conference on Computer Vision}, pages 14034--14044, 2023.

\bibitem[Denninger et~al.(2023)Denninger, Winkelbauer, Sundermeyer, Boerdijk, Knauer, Strobl, Humt, and Triebel]{Denninger2023}
M.~Denninger, D.~Winkelbauer, M.~Sundermeyer, W.~Boerdijk, M.~Knauer, K.~H. Strobl, M.~Humt, and R.~Triebel.
\newblock Blenderproc2: A procedural pipeline for photorealistic rendering.
\newblock \emph{Journal of Open Source Software}, 8\penalty0 (82):\penalty0 4901, 2023.
\newblock \doi{10.21105/joss.04901}.
\newblock URL \url{https://doi.org/10.21105/joss.04901}.

\bibitem[Wang and Olson(2016)]{wang2016apriltag}
J.~Wang and E.~Olson.
\newblock Apriltag 2: Efficient and robust fiducial detection.
\newblock In \emph{2016 IEEE/RSJ International Conference on Intelligent Robots and Systems (IROS)}, pages 4193--4198, 2016.
\newblock \doi{10.1109/IROS.2016.7759617}.

\bibitem[Schönberger and Frahm(2016)]{schonberger2016sfm}
J.~L. Schönberger and J.-M. Frahm.
\newblock Structure-from-motion revisited.
\newblock In \emph{2016 IEEE Conference on Computer Vision and Pattern Recognition (CVPR)}, pages 4104--4113, 2016.
\newblock \doi{10.1109/CVPR.2016.445}.

\bibitem[Contributors(2022)]{xrsfm}
X.~Contributors.
\newblock Openxrlab structure-from-motion toolbox and benchmark.
\newblock \url{https://github.com/openxrlab/xrsfm}, 2022.

\bibitem[Yu et~al.(2024)Yu, Sattler, and Geiger]{Yu2024GOF}
Z.~Yu, T.~Sattler, and A.~Geiger.
\newblock Gaussian opacity fields: Efficient adaptive surface reconstruction in unbounded scenes.
\newblock \emph{ACM Transactions on Graphics}, 2024.

\bibitem[Kirillov et~al.(2023)Kirillov, Mintun, Ravi, Mao, Rolland, Gustafson, Xiao, Whitehead, Berg, Lo, Doll{\'a}r, and Girshick]{kirillov2023segany}
A.~Kirillov, E.~Mintun, N.~Ravi, H.~Mao, C.~Rolland, L.~Gustafson, T.~Xiao, S.~Whitehead, A.~C. Berg, W.-Y. Lo, P.~Doll{\'a}r, and R.~Girshick.
\newblock Segment anything.
\newblock \emph{arXiv:2304.02643}, 2023.

\end{thebibliography}
\clearpage

% \documentclass{article}
% \usepackage{amssymb}
% \usepackage{verbatim}
% \usepackage{graphicx} 
% \usepackage{booktabs}  
% \usepackage{multirow}   
% \usepackage{array}      
% \usepackage[table]{xcolor}
% \usepackage{pdflscape}
% \usepackage{amsmath}

% % \usepackage{corl_2025} % Use this for the initial submission.
% % \usepackage[]{corl_2025} % Uncomment for the camera-ready ``final'' version.
% \usepackage[preprint]{corl_2025} % Uncomment for pre-prints (e.g., arxiv); This is like ``final'', but will remove the CORL footnote.
% \appendix
% \title{One View, Many Worlds: Single-Image to 3D Object Meets Generative Domain Randomization for One-Shot 6D Pose Estimation}

% % The \author macro works with any number of authors. There are two
% % commands used to separate the names and addresses of multiple
% % authors: \And and \AND.
% %
% % Using \And between authors leaves it to LaTeX to determine where to
% % break the lines. Using \AND forces a line break at that point. So,
% % if LaTeX puts 3 of 4 authors names on the first line, and the last
% % on the second line, try using \AND instead of \And before the third
% % author name.

% % NOTE: authors will be visible only in the camera-ready and preprint versions (i.e., when using the option 'final' or 'preprint'). 
% % 	For the initial submission the authors will be anonymized.

% \begin{document}
% \maketitle
\appendix
\setcounter{section}{0}
\section*{Appendix}
\label{sec:appendix}
    This appendix is organized as follows. In section A, we describe how carefully crafted text prompts guide the generation of diverse 3D models using Trellis~\cite{xiang2024trellis}, emphasizing the impact of linguistic variation on visual diversity. Section B details the construction of our synthetic training dataset, including the scene setup and statistical distributions across object pose, visibility, and distance. In section C, we report additional experimental results, including per-dataset performance analysis, failure mode characterization, and discussions of robustness across challenging conditions. In section D introduces a complete pipeline for constructing a ground-truth dataset for unseen objects in real-world scenes, encompassing scene acquisition, 3D reconstruction, coordinate alignment, and mask generation. Together, these components provide comprehensive support for evaluating our method in both synthetic and real-world conditions.

\section{Text Prompts and Model Output Diversity}
\label{diversity}
    As described in Section 3.5, we employ Trellis~\cite{xiang2024trellis} to generate diverse textures. To this end, we use a prompt to instruct the VLM to produce suitable input prompts for Trellis:

    Prompt: Given the image of a [OBJECT], generate a detailed and realistic prompt for a 3D modeling system to create diverse variants of this object. The prompt should request:
    A series of unique but plausible 3D models.
    Variations in design style (e.g., minimalist, industrial, futuristic, ergonomic).
    Inclusion of all essential functional components visible in the image.
    Use of realistic materials (e.g., matte plastic, brushed metal) with subtle imperfections for authenticity.
    Distinct color schemes with aesthetic and functional considerations (e.g., color-coded controls).
    Practicality and usability in real-world scenarios.
    Format the output as a natural-language instruction starting with "Generate...".
    
\begin{figure}[htbp]
\centerline{
\includegraphics[width=1\columnwidth]{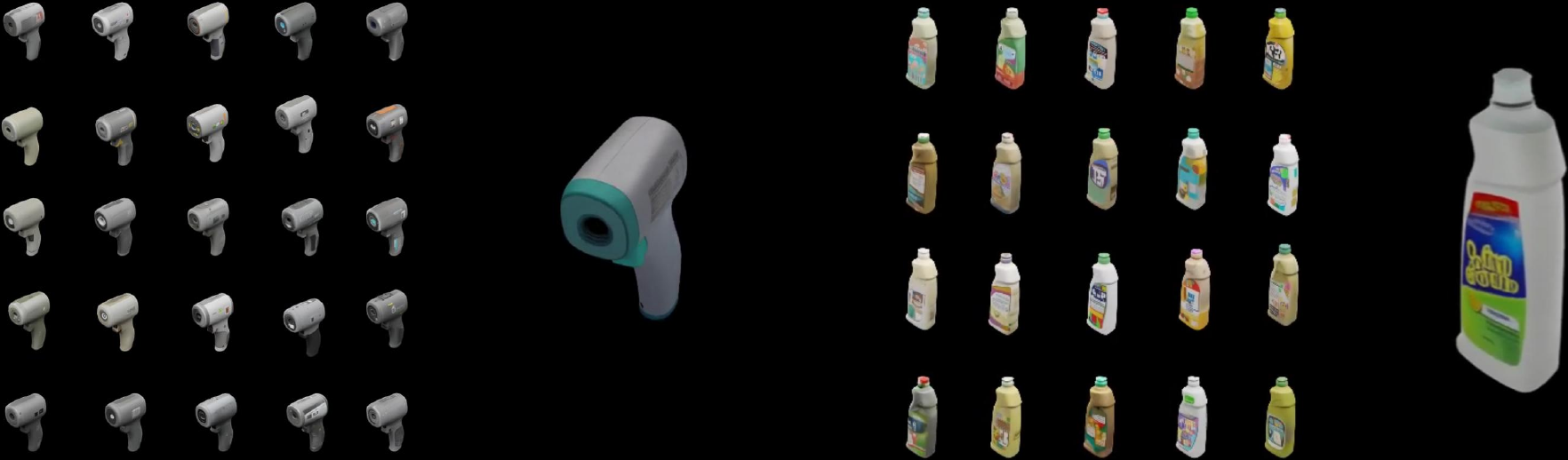}
}

\caption{
\textbf{Diversified Models. }This figure showcases the results of two types of diversified texture generation. For each object, the original model is displayed on the right, while the model with diversified textures is shown on the left. It can be observed that our proposed method, which utilizes text prompts to promote texture diversity, is capable of generating rich and varied surface textures. This contributes to the enhancement and diversification of training dataset generation.
}

\label{fig:models}
\end{figure}

    We then input the models generated based on an anchor image into our pipeline, resulting in the models shown in Fig.~\ref{fig:models}. It is evident that the generated models exhibit a rich diversity of textures, which helps further narrow the domain gap between the training dataset and real-world objects in the subsequent step of dataset generation. The experimental results validate the effectiveness of this approach, as discussed in the main text.

\section{Statistical Distribution of the Training Dataset}
\begin{figure}[htbp]
\centerline{
\includegraphics[width=1\columnwidth]{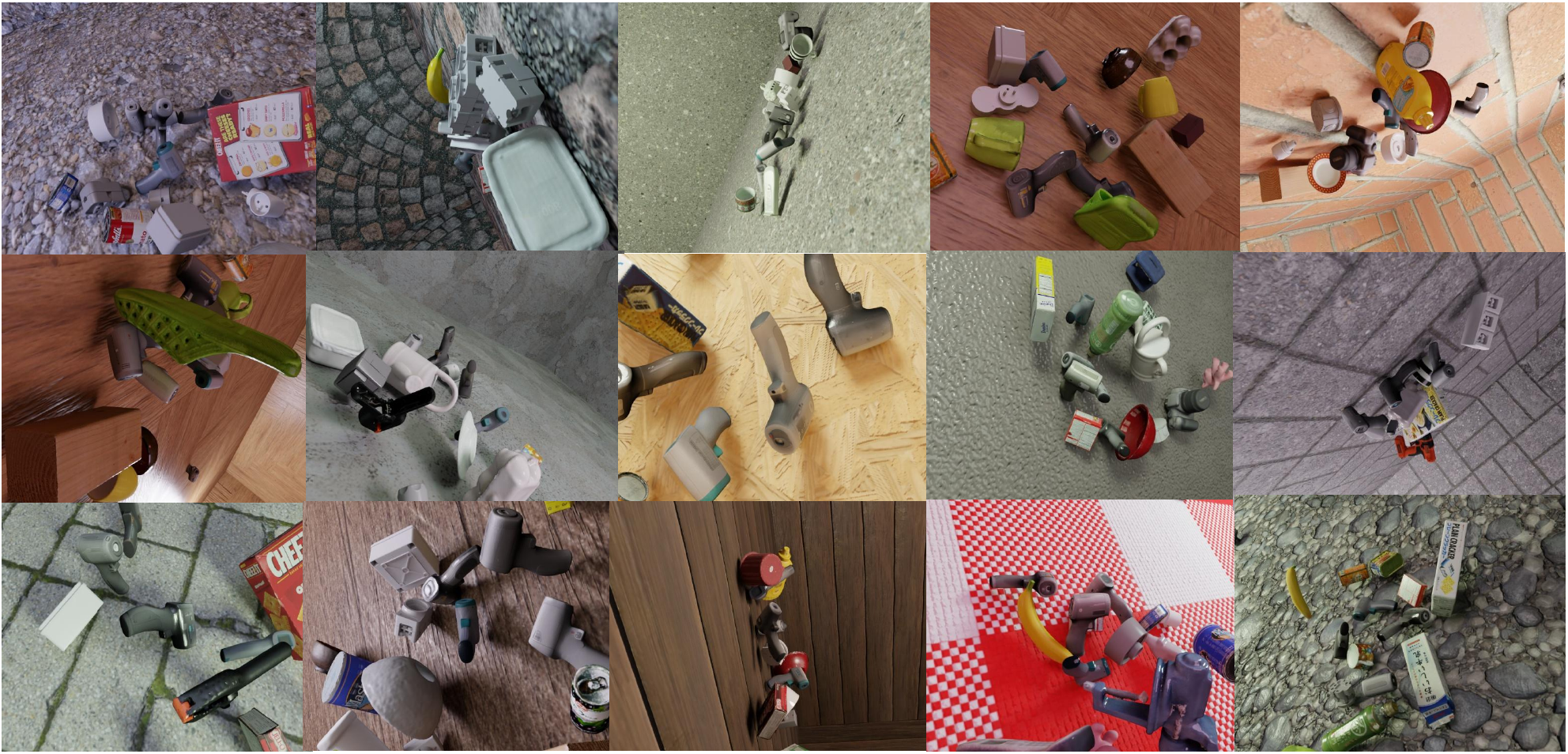}
}

\caption{
\textbf{Generated training dataset. }This figure illustrates the dataset generated using our diversified texture models. It is evident that our generated dataset encompasses a rich variety of backgrounds, object poses, occlusion relationships, and lighting conditions. This contributes to narrowing the domain gap between the training dataset and real-world scenarios.
}

\label{fig:train_dataset}
\end{figure}
\paragraph{The method for constructing a training dataset. }
    First, we apply the method described in Sec. 3.5 to perform diversified texture generation on existing object models using text prompts, thereby creating 100 differently textured models as shown in Fig.~\ref{fig:models}, These models are then fed into the BlenderProc~\cite{Denninger2023} rendering pipeline.
    For each synthetic scene, an initial model and three randomly selected diversified texture models are chosen as training targets; simultaneously, ten objects are randomly picked from the BOP dataset~\cite{hodan2018bop} to act as occluders, and a background environment is constructed by selecting a random texture map from the CCTextures.
    Subsequently, 100 camera positions are randomly determined with each camera oriented towards the geometric center of these 14 objects (4 targets + 10 occluders), incorporating eccentric noise into the camera positions while also introducing random perturbations to the rotation angles around their Z-axis to simulate variations in real-world shooting conditions.
    Through this process, we have constructed a large-scale, highly varied synthetic training dataset. Fig.~\ref{fig:train_dataset} provides examples from this dataset, illustrating its rich variety of lighting conditions, degrees of occlusion, and object scales (distances from the camera), significantly enhancing the robustness and accuracy of the model in pose estimation tasks, as discussed in the main text.

\begin{figure}[htbp]
\centerline{
\includegraphics[width=0.8\columnwidth]{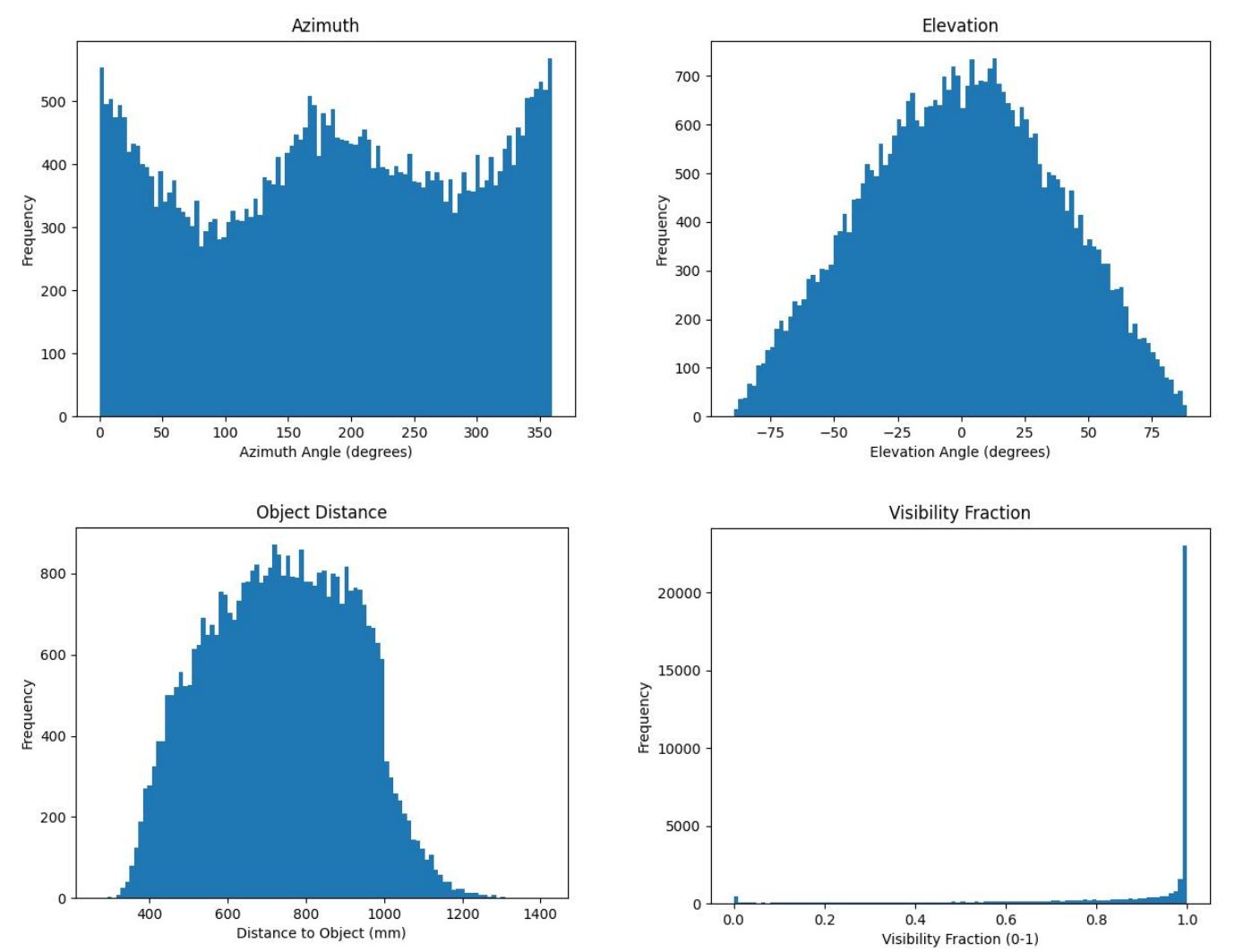}
}

\caption{
\textbf{Distribution Analysis of Training Dataset. } Top-left: Distribution of Azimuth angles for objects within the generated dataset;
Top-right: Distribution of Elevation angles for objects;
Bottom-left: Distribution of object distances from the camera;
Bottom-right: Distribution of the proportion of object area visibility;
These plots demonstrate that the constructed training dataset encompasses a rich diversity of poses and occlusion scenarios, underpinning its comprehensive variability and realism.
}

\label{fig:train_dataset_distribution}
\end{figure}
\paragraph{The object pose distribution of the training dataset.}
    We present the distributions of Azimuth, Elevation, Object Distance, and Visibility in the final generated dataset, as shown in Fig.~\ref{fig:train_dataset_distribution}. From these distributions, it is evident that our dataset exhibits a high degree of diversity in both azimuth and elevation angles, covering a wide range of horizontal and vertical orientations of objects relative to the camera.
    In terms of object distance, the dataset includes samples ranging from close-up to far-field scenarios, simulating the variation in depth at which objects may appear in real-world settings. Regarding visibility, due to the inclusion of occluders and randomized scene layouts, the extent of visible object regions varies significantly across images, resulting in diverse levels of occlusion. 
    Taken together, the multi-dimensional diversity in our dataset closely mimics the complexity of real-world camera-captured scenes. This design choice effectively enhances the generalization capability and robustness of model training.

\section{Fine-tuning Details}
\label{fitune_details}
\begin{figure}[htbp]
\vspace{-0.4cm} 
\centerline{
\includegraphics[width=0.8\columnwidth]{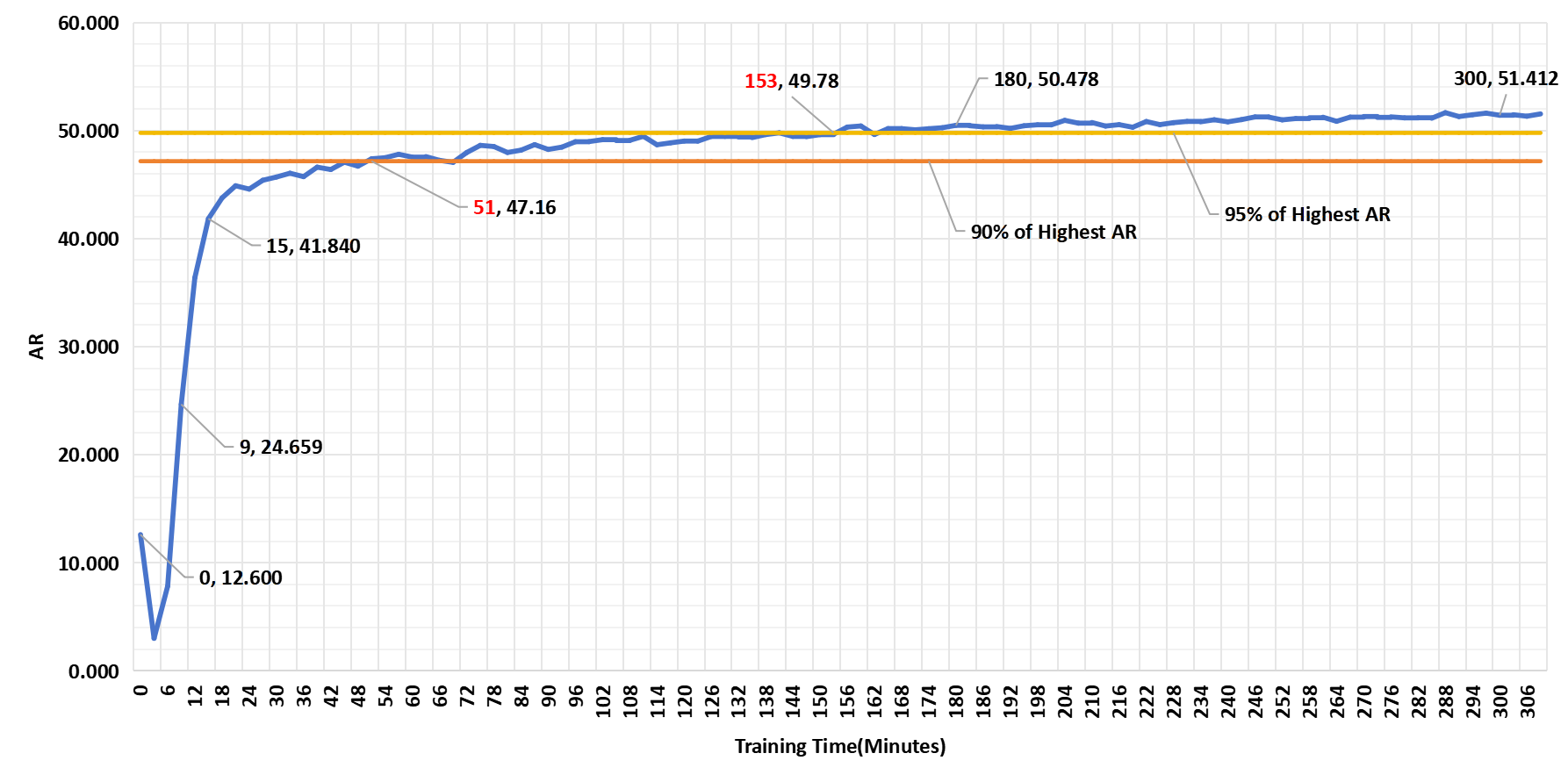}
}
\vspace{-0.45cm} 
\caption{
\textbf{Finetune Performance with Time. }
}

\label{fig:finetune_time}
\vspace{-0.3cm} 
\end{figure}

When comparing with other one-shot methods, we did not fine-tune our model to ensure a fair comparison, making the evaluation valid under the same settings. In cases of object with long tail distribution where fine-tuning is desired, generating a diversified models takes around 10 minutes on an A800 GPU, while generating the training dataset requires approximately 10 minutes on an L20 GPU. As shown in Fig.~\ref{fig:finetune_time}, the fine-tuning stage, with training a single LoRA module, takes about 51 minutes to achieve 90\% of highest AR on an single L20 GPU with 48GB of memory. Compared to redesigning the method from scratch, this fine-tuning process is computationally efficient and the time cost is acceptable in practice.

The detailed information of all modules in One-2-3-Pose is summarized in Table~\ref{tab:module_properties}. Specifically, the Hi3DGen module is fine-tuned by concatenating the texture features with the original feature vectors. The Pose Generation module of FoundationPose is fine-tuned using the domain randomization strategy described in Section~3.5.

\begin{table}[htbp]
\vspace{-0.4cm} 
\tiny
\centering
\caption{Overview of modules and their properties.}
\vspace{-0.1cm} 
\label{tab:module_properties}
\begin{tabular}{m{1.5cm} m{3.3cm} m{2.2cm} c}
\toprule
\textbf{Module Name} & \textbf{Input} & \textbf{Output} & \textbf{Is Fine-tuned?} \\ 
\midrule 
Hi3DGen              & Segmented anchor image & Normalized 3D mesh & \cmark \\
SuperPoint           & RGB templates and segmented anchor image & Detected 2D keypoints with descriptors & \xmark \\
SuperGlue            & Extracted 2D keypoints and descriptors & Matched 2D-2D correspondences & \xmark \\
PnP                  & Matched 2D-3D point correspondences and camera intrinsics & Estimated object pose (rotation $R$ and translation $t$) & \xmark \\
Scale Optimization & 3D model points in object coordinates and corresponding camera-space points & Scale factor $s$ & \xmark \\
FoundationPose     & Anchor image with segmentation mask and template under coarse pose & Refined object pose & \cmark \\
Trellis            & Normalized 3D mesh and text-based prompt for texture generation & Texture-diversified 3D model & \xmark \\
\bottomrule
\end{tabular}
\vspace{-0.3cm} 
\end{table}
\begin{table}[htbp]
\centering
\caption{Processing Stages and Time Consumption}
\label{tab:processing_stages}
\begin{tabular}{lcc}
\toprule
\textbf{Stage}        & \textbf{Submodule}         & \textbf{Time Cost (s)} \\ 
\midrule
Model Generation      & Model Generation           & 9.10                     \\
                      & Model Export               & 12.84                     \\
\addlinespace
Scale Recovery        & Coarse Alignment       & 0.75                     \\
(1 Round)             & Fine Alignment         & 0.23                     \\
\addlinespace
Pose Estimation       & Pose Generation           & 0.88                    \\
(FoundationPose)      & Pose Selection          & 0.42                      \\
\bottomrule
\end{tabular}
\end{table}

\section{Additional experiments}
\label{additional_exp}

\paragraph{Metrics. }
    We assess performance using the metrics defined by the BOP challenge, specifically focusing on Average Recall (AR) for Visual Surface Discrepancy (VSD), Maximum Symmetry-aware Surface Distance (MSSD), and Maximum Symmetry-aware Projection Distance (MSPD)~\cite{hodan2018bop}. These metrics offer complementary insights into pose accuracy by evaluating recall rates across various thresholds. This comprehensive evaluation approach ensures a thorough assessment of algorithmic performance in diverse scenarios, reflecting different aspects of precision and robustness in pose estimation.
    
\begin{figure}[ht]
\centerline{
\includegraphics[width=1\columnwidth]{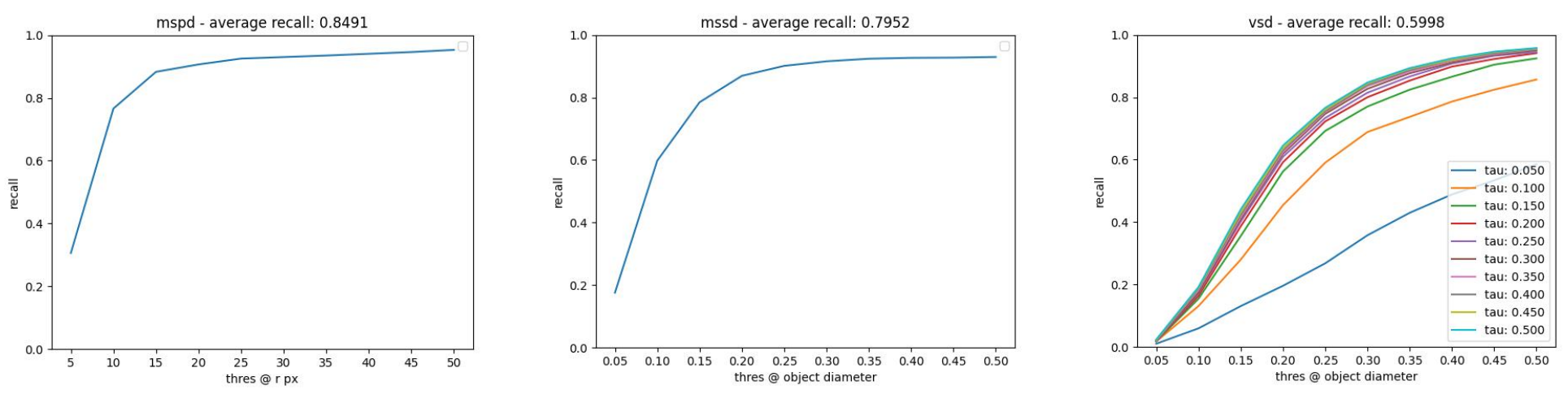}
}

\caption{
\textbf{Visualization of Metrics on the LM-O Dataset. }Left: Illustration of the MSPD metric variation with respect to the visible size of objects in pixels.
Middle: Demonstration of the MSSD metric variation according to the object size in meters.
Right: Presentation of the VSD metric variation with respect to $\tau$ (denoting the misalignment tolerance) and object size.
These graphs effectively capture the metrics' dependencies on object visibility, size, and alignment tolerance, providing insights into their varying influences under different conditions.
}

\label{fig:lmo_details_figure}
\end{figure}

\begin{figure}[ht]
\centerline{
\includegraphics[width=0.6\columnwidth]{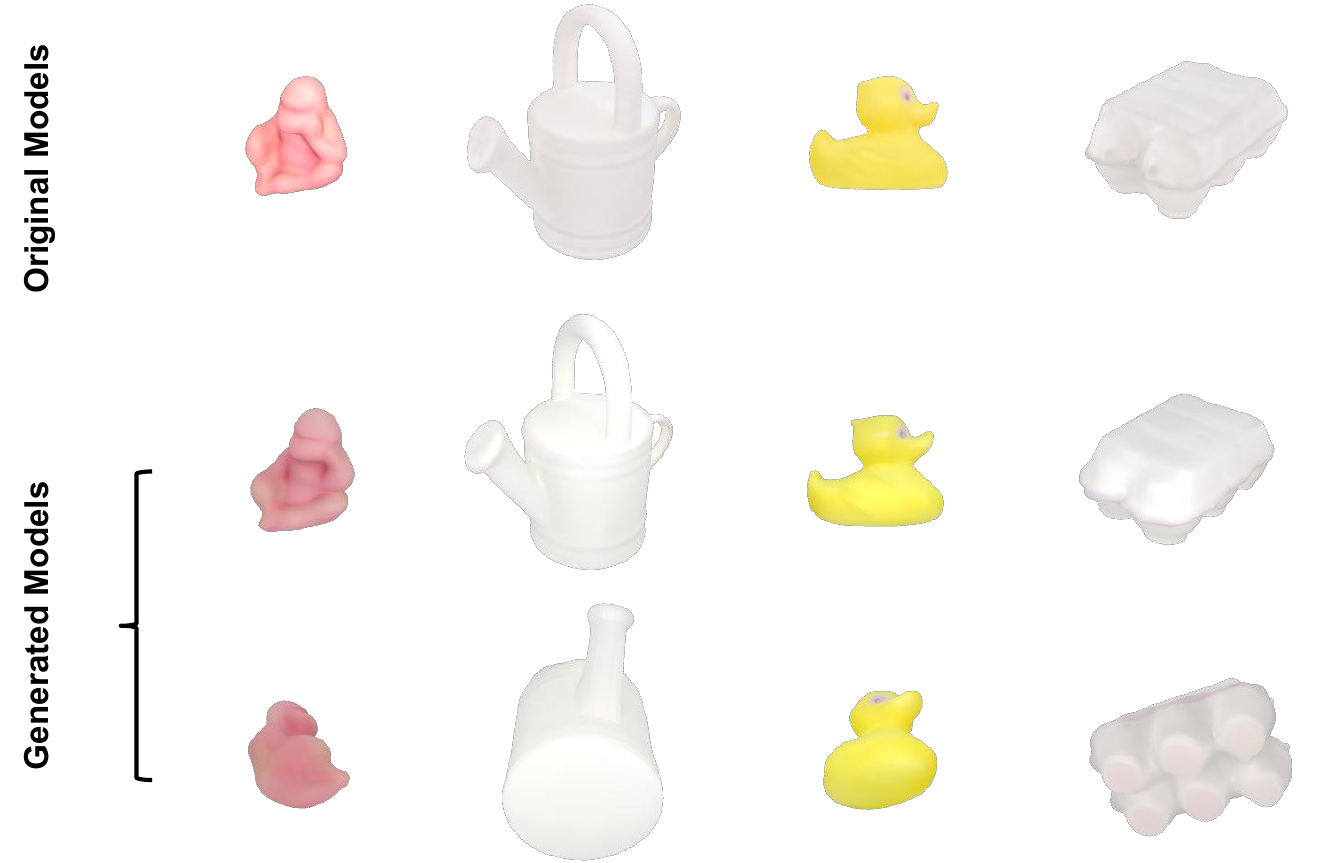}
}

\caption{
\textbf{Comparison Between Original and Generated Models on LM-O dataset. }The first row displays the original object models, while the second row shows the generated models under the same pose. The third row presents a bottom-view comparison of the generated models. As can be seen, the generated models exhibit high quality and closely resemble the original objects in terms of texture and structure, demonstrating the effectiveness of our generation and scale recovery approach.
}

\label{fig:LMO_models}
\end{figure}

\paragraph{Performance Analysis on the LM-O Dataset. }
    The LM-O dataset~\cite{brachmann2014learning} comprises 12 objects, predominantly distinguished by their lack of texture and frequent occlusion. Performance metrics for each object are summarized in Table~\ref{tab:metrics_lm-o}, where $\tau$ denotes the misalignment tolerance. As shown in Fig.~\ref{fig:lmo_details_figure}, performance degrades when the target object occupies a small region in the image. Nonetheless, the method still achieves relatively robust results under such challenging conditions.
    As observed in Table~\ref{tab:metrics_lm-o}, the model exhibits stable pose estimation across most categories when the translational error threshold exceeds 0.15, indicating consistent precision and reliability in 6D pose estimation. However, the ``ape'' category shows notably lower accuracy, primarily due to its minimal textural information and ambiguous geometry. These characteristics result in a mismatch between the reconstructed and ground-truth models, which in turn hampers accurate model alignment and pose estimation.
    Despite this limitation, as illustrated in Fig.~\ref{fig:LMO_models}, the proposed method demonstrates strong and consistent performance across the majority of object categories, highlighting its generalization capability under diverse shape and textural conditions in real-world 6D object pose estimation tasks.
    \begin{table}[ht]
\centering
\caption{Detailed Metrics on the LM-O Dataset.}
\label{tab:metrics_lm-o}
\begin{tabular}{l c c c c c c c c c c}
\toprule
\multirow{2}{*}{Metrics} & \multicolumn{9}{c}{object} & \multirow{2
}{*}{avg} \\
\cmidrule(lr){2-10
}
& ape & can & cat & driller & duck & eggbox & glue & holepunch & & \\
\midrule
MSPD & 79.6 & 93.7 & 80.8 & 82.1 & 84.4 & 86.0 & 77.9 & 91.5 & & 84.9
 \\
MSSD & 64.8 & 92.6 & 71.2 & 85.8 & 72.9 & 80.1 & 75.8 & 87.9 & & 79.6
 \\
VSD & 37.5 & 75.5 & 57.7 & 71.2 & 66.1 & 56.6 & 51.9 & 58.2 & & 60.0
 \\
AR & 60.6 & 87.3 & 69.9 & 79.7 & 74.5 & 74.2 & 68.5 & 79.2 & & 74.8
 \\
\bottomrule
\end{tabular}
\end{table}
    \begin{figure}[ht]
\centerline{
\includegraphics[width=1\columnwidth]{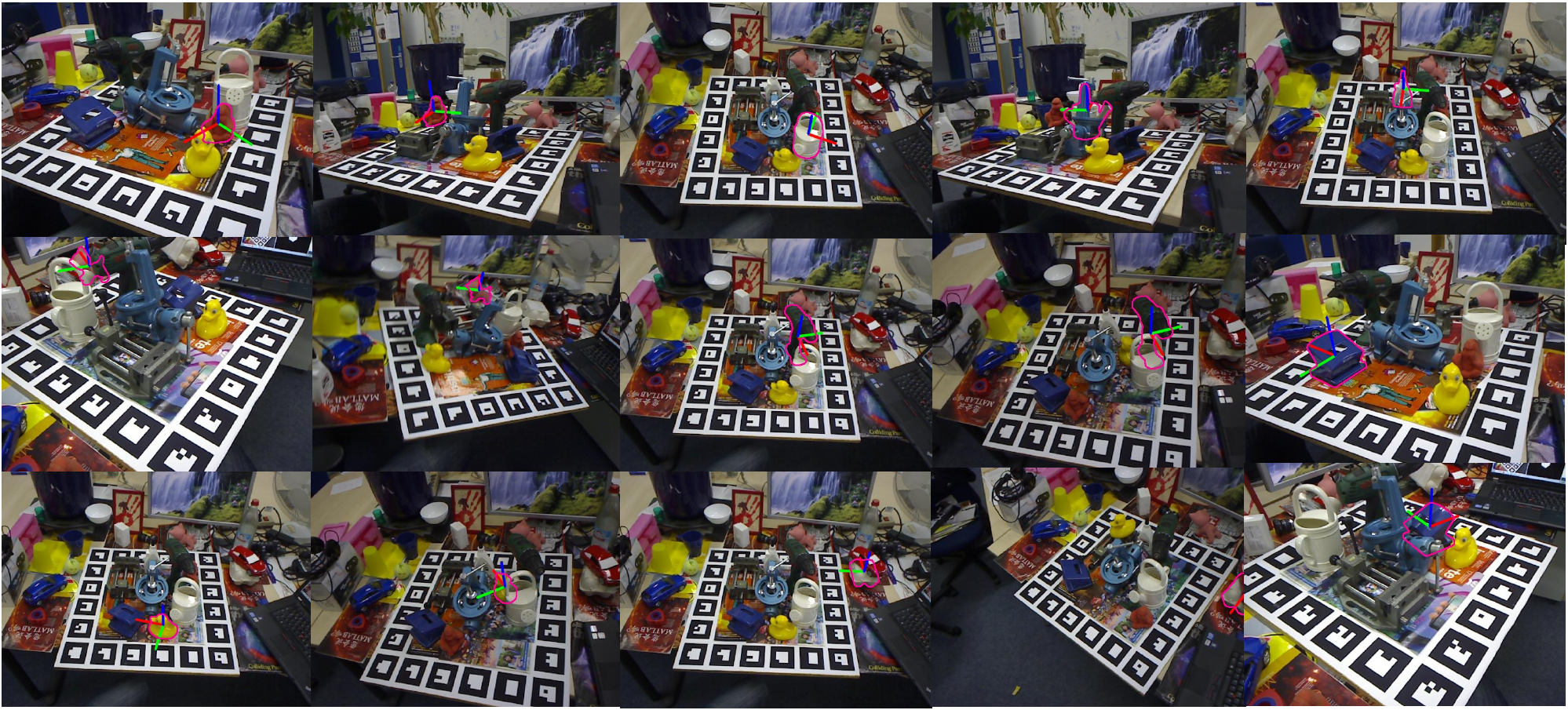}
}

\caption{
\textbf{Performance in LM-O dataset. } In each image, the \textcolor{red}{red}, \textcolor{green}{green}, and \textcolor{blue}{blue} lines represent the x, y, and z axes of the model, respectively, while the \textcolor{magenta}{pink} line shows the rendered contour under the estimated pose. By comparing the rendered contour with the ground-truth outline of the object, it is evident that our method is highly robust, performing well across various objects and under different occlusion scenarios.
}

\label{fig:LMO_performance}
\end{figure}
    
\begin{figure}[ht]
\centerline{
\includegraphics[width=1\columnwidth]{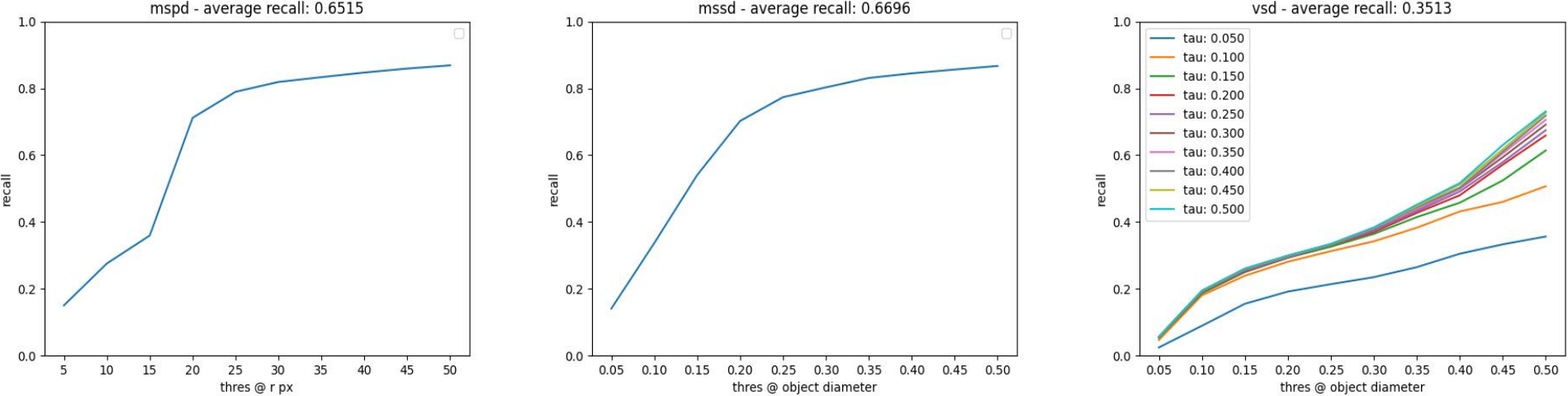}
}

\caption{
\textbf{Visualization of Metrics on the TYOL Dataset. }Left: Illustration of the MSPD metric variation with respect to the visible size of objects in pixels.
Middle: Demonstration of the MSSD metric variation according to the object size in meters.
Right: Presentation of the VSD metric variation with respect to $\tau$ (denoting the misalignment tolerance) and object size.
These graphs effectively capture the metrics' dependencies on object visibility, size, and alignment tolerance, providing insights into their varying influences under different conditions.
}

\label{fig:tyol_details_figure}
\end{figure}

\paragraph{Performance Analysis on the TOYL Dataset. }
    The TOYL~\cite{hodan2018bop} test dataset contains 21 objects, primarily distinguished by complex lighting conditions and the fact that most objects are positioned relatively far from the camera. Performance metrics for each object are summarized in Table~\ref{tab:metrics_tyol}. As illustrated in Fig.~\ref{fig:tyol_details_figure}, performance declines when the target occupies only a small portion of the image, which is often due to the considerable distance from the camera. This makes texture the primary discriminative cue for pose estimation, as the objects' high symmetry reduces the effectiveness of other features.
    Despite these challenges, the model exhibits relatively stable performance across most object categories when the translational error threshold exceeds 0.15, indicating consistent behavior under varying conditions. As shown in Table~\ref{tab:metrics_tyol}, the method faces notable difficulties with objects numbered 04 and 18. Their high symmetry restricts reliable orientation estimation to surface texture cues alone, often leading to mismatches when the opposite side of the object is visible.
    Nevertheless, for the majority of other objects, the proposed method achieves stable and accurate pose estimation, demonstrating strong robustness and adaptability even when objects are positioned at significant distances from the camera.

    \begin{table}[ht]
    \tiny
    \centering
    \caption{Detailed  Metrics on the TOYL Dataset.}
    \label{tab:metrics_tyol}
    \resizebox{1\columnwidth}{!}{
        \begin{tabular}{l c c c c c c c c c c c c c c c c c c c c c c}
        \toprule
        \multirow{2}{*}{Metrics} & \multicolumn{21}{c}{object} & \multirow{2}{*}{avg} \\
        \cmidrule(lr){2-22}
        & 01 & 02 & 03 & 04 & 05 & 06 & 07 & 08 & 09 & 10 & 11 & 12 & 13 & 14 & 15 & 16 & 17 & 18 & 19 & 20 & 21 & \\
        \midrule
        MSPD & 80.1 & 48.5 & 50.5 & 36.8 & 72.1 & 46.9 & 74.2 & 42.2 & 85.1 & 82.2 & 77.8 & 84.6 & 83.1 & 68.9 & 80 & 72.5 & 59.5 & 31.9 & 74.4 & 65.3 & 54.5 & 55.7 \\
        MSSD & 76.4 & 51.1 & 40.8 & 33.6 & 75.9 & 47.5 & 73 & 63.3 & 91.6 & 83.1 & 76.1 & 84.7 & 86.1 & 79.8 & 88.5 & 77.7 & 48.3 & 26 & 90.3 & 61.8 & 54.1 & 65.1 \\
        VSD & 31.1 & 33.6 & 41.7 & 18.5 & 25.5 & 35 & 31.4 & 24.9 & 42.8 & 38.1 & 25.6 & 45.6 & 47.5 & 52.8 & 51.3 & 44.3 & 40.5 & 26.7 & 34.4 & 27.3 & 20.1 & 67.0 \\
        AR & 62.5 & 44.4 & 44.3 & 29.6 & 57.8 & 43.1 & 59.5 & 43.5 & 73.2 & 67.8 & 59.8 & 71.6 & 72.2 & 67.2 & 73.3 & 64.8 & 49.4 & 28.2 & 66.4 & 51.5 & 42.9 & 35.2 \\
        \bottomrule
        \end{tabular}
    }
\end{table} 
    \begin{figure}[ht]
\centerline{
\includegraphics[width=1\columnwidth]{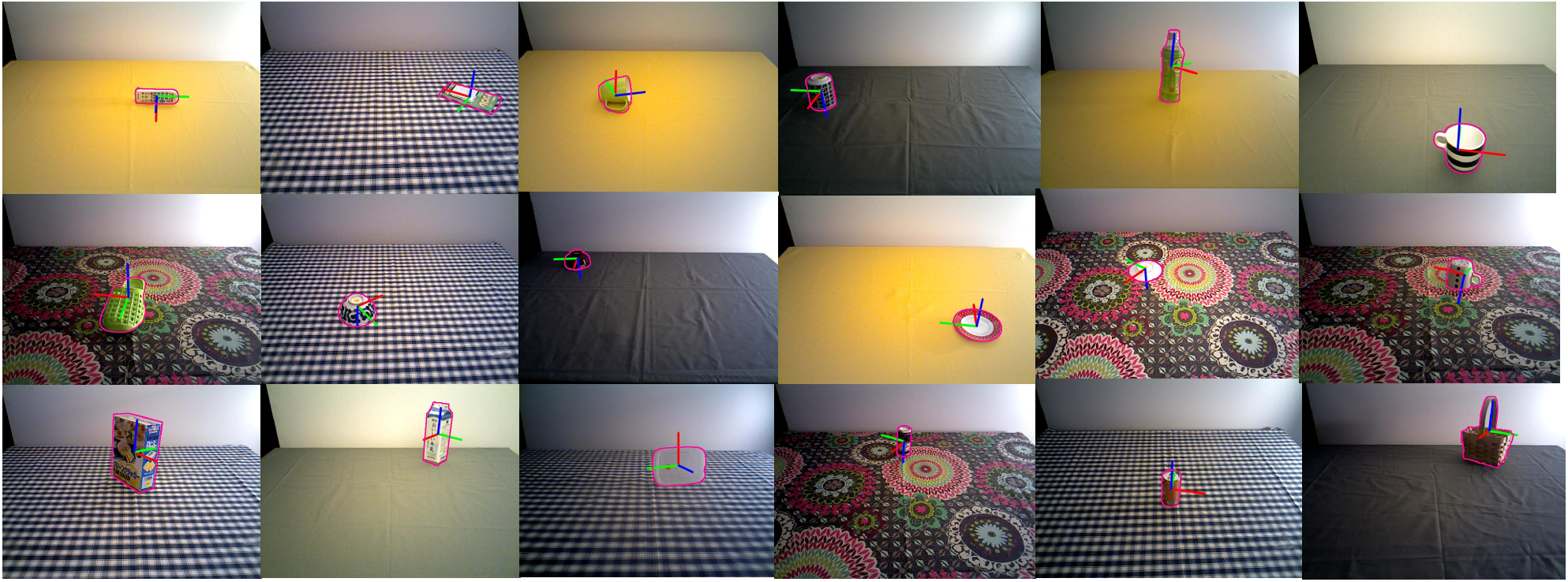}
}

\caption{
\textbf{Performance in TYOL dataset. } In each image, the \textcolor{red}{red}, \textcolor{green}{green}, and \textcolor{blue}{blue} lines represent the x, y, and z axes of the model, respectively, while the \textcolor{magenta}{pink} line shows the rendered contour under the estimated pose. By comparing the rendered contour with the ground-truth outline of the object, it is evident that our method is highly robust, performing well across various objects and under different occlusion scenarios.
}

\label{fig:TYOL_performance}
\end{figure}
    \begin{figure}[ht]
\centerline{
\includegraphics[width=1\columnwidth]{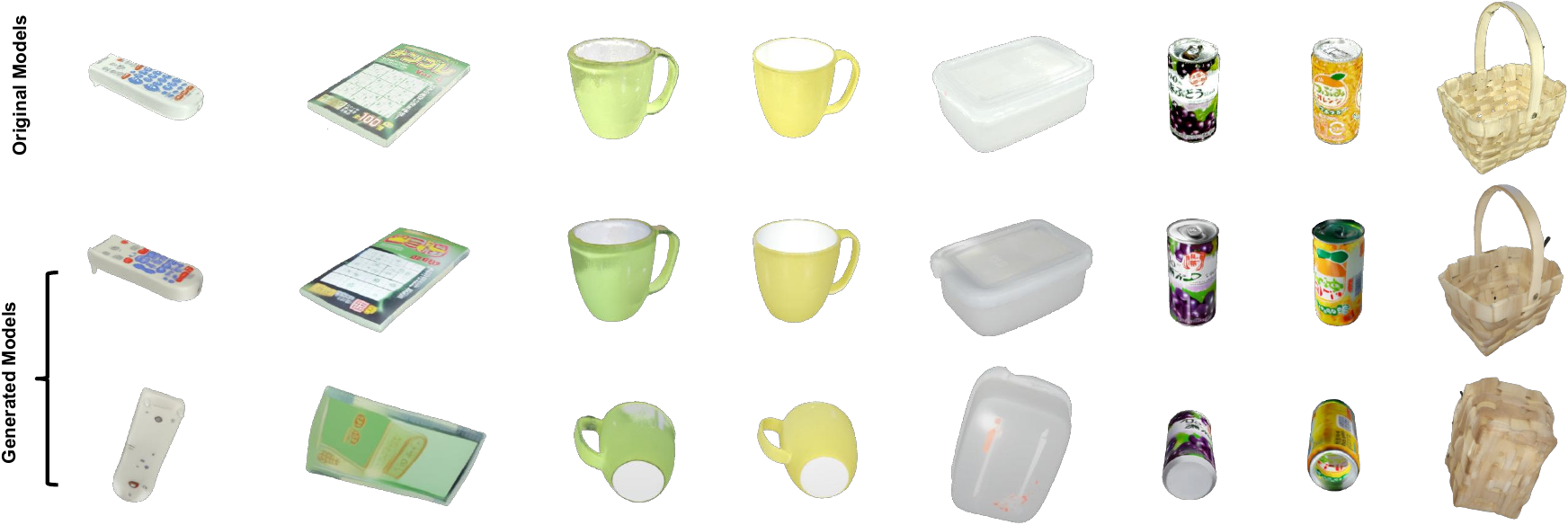}
}

\caption{\textbf{Comparison Between Original and Generated Models on TOYL dataset. }The first row displays the original object models, while the second row shows the generated models under the same pose. The third row presents a bottom-view comparison of the generated models. As can be seen, the generated models exhibit high quality and closely resemble the original objects in terms of texture and structure, demonstrating the effectiveness of our generation and scale recovery approach in TOYL dataset.
}

\label{fig:TYOL_models}
\end{figure}

\begin{figure}[htbp]
\centerline{
\includegraphics[width=0.8\columnwidth]{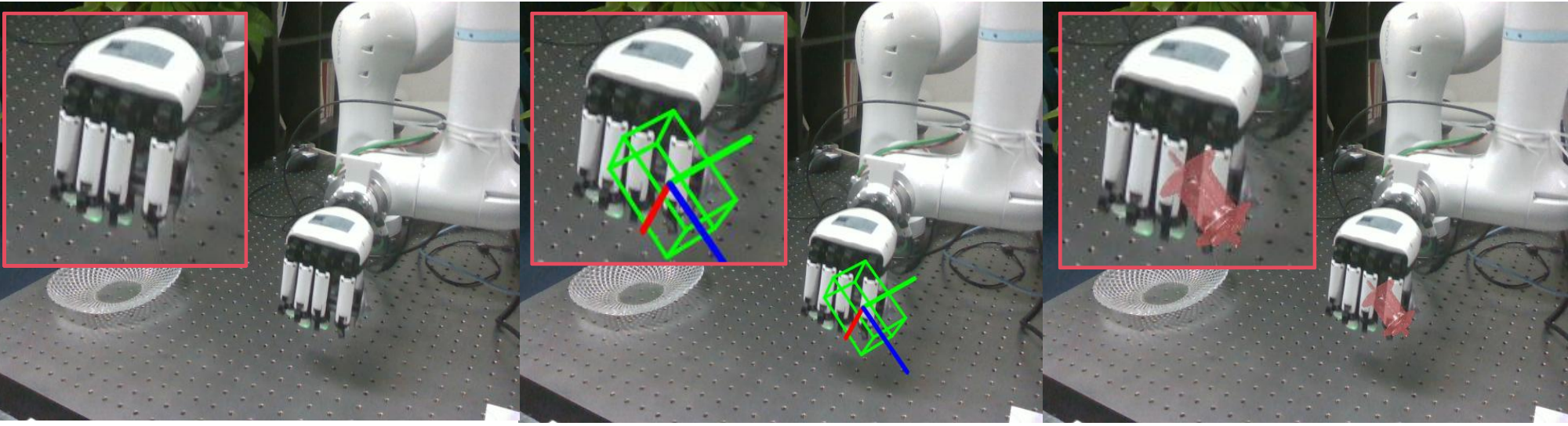}
}

\caption{
\textbf{Failure modes in real-world test. } Left: The original image.
Middle: Detection results showing the bounding box and model's XYZ axes.
Right: Rendered model image based on the detected pose.
This visualization demonstrates that our method struggles in scenarios characterized by high symmetry, significant occlusion, and severe motion blur. Despite these challenges, the method shows promise in more favorable conditions, highlighting areas for potential improvement.
}

\label{fig:failure_modes}
\end{figure}

\begin{figure}[htbp]
\centerline{
\includegraphics[width=0.8\columnwidth]{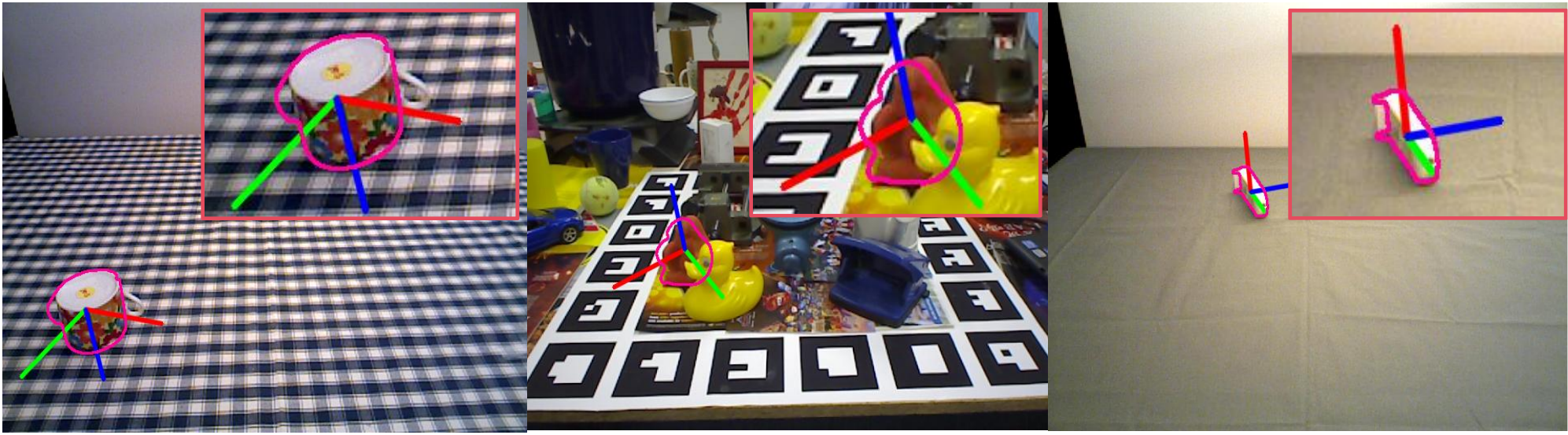}
}

\caption{
\textbf{Failure modes in LMO and TOYL dataset. } In each image, the \textcolor{red}{red}, \textcolor{green}{green}, and \textcolor{blue}{blue} lines represent the x, y, and z axes of the model, respectively, while the \textcolor{magenta}{pink} line shows the rendered contour under the estimated pose. The top-right corner of each image displays an enlarged view of the corresponding failure region.
}

\label{fig:dataset_failure_modes}
\end{figure}

\paragraph{Failure modes. }
    As shown in Fig.~\ref{fig:failure_modes} and Fig.~\ref{fig:dataset_failure_modes}, the proposed method exhibits performance limitations in scenarios involving severe occlusion, strong motion blur, as well as objects with low texture and high symmetry.

\paragraph{More details about the real-world experiments. }
    The primary perception system for both tasks relied on Intel RealSense cameras, providing RGB and depth information at a resolution of $640\times480$ and a frame rate of 30 $Hz$. For \textbf{Task 1} (ROKAE Pick-and-Place), the camera was fixed next to the robotic arm, approximately 0.5 meter above the center of the manipulation area, angled at 15 degrees from the vertical. For \textbf{Task 2} (Dual-Arm AgileX PiPER Manipulation), an eye-in-hand camera on each arm, offers close-up views beneficial for fine manipulation, whereas a fixed external camera provides a broader view. All cameras were calibrated prior to the experiments using standard procedures to ensure accurate spatial information.

    \textbf{Task 1: Pick-and-Place with ROKAE Arm and XHAND1}~~This task evaluated the ability of a ROKAE collaborative robot arm equipped with an XHAND1 dexterous hand to pick up a variety of objects from a starting location and place them accurately at a designated target pose.
    
    \textbf{1. Hardware Configuration.}~~
    \begin{enumerate}
        \item ROKAE Robot Arm ($\times1$), is designed for collaborative applications, featuring a cabinet-free design and integrated torque sensors in each joint, promoting deployment flexibility and safety, with 6-DoF.
        \item XHAND1 Dexterous Hand ($\times1$), has 12 active degrees of freedom and supports force control with haptic sensor feedback, allowing us to adapt to object shapes and ensure stable grasps. The hand is equipped with five $270^{\circ}$ three-dimensional encircling tactile array sensors on the fingertips, providing a tactile resolution of $12\times10$ per fingertip and sensing 3D forces including tangential components (X and Y). Joint sensors provide position, velocity, temperature, and current (torque) information.
    \end{enumerate}
    The XHAND1 was mechanically attached to the ROKAE arm's end-effector flange.
    
    \textbf{2. Task Protocol.}~~
    A set of 15 diverse objects was used to evaluate the pick-and-place capabilities. Objects were presented randomized within a $10\times 10 cm^2$ area with orientation randomized $\pm 15$ degrees around the vertical axis. The target placement pose for each object was a plate into which the object needed to be placed. 

    For the 12-DoF XHAND1, grasp planning involved a pre-defined grasp synergies. Tactile and force feedback from the XHAND1's sensors were used to confirm contact, and monitor grasp stability during lift and transport, leveraging its force-position control capabilities.

   An initial 6D pose estimate of the object was obtained from the perception system (detailed in the main paper). This pose was used to determine the robot's approach trajectory to the object and plan the motion to the target placement location. The system operated on the assumption that this initial pose was sufficiently accurate for grasp execution. Then we use the pose tracking methods to guide the motion of the arm in real-time.

   For each of the 15 objects, 30 pick-and-place trials were conducted. Between trials, the object was randomized within the defined start region, and the scene was reset.

    \textbf{3. Success Criteria.}~~

    \begin{enumerate}
        \item Successful Grasp: object was securely held by the XHAND1, lifted at least 5 cm clear of the support surface, and maintained a stable grasp without slipping or being dropped during the initial lift.

        \item Successful Transport: The object was moved from the pick location to the designated target area without any collisions with the environment or being dropped.

        \item Successful Placement: The object was released at the target pose, the object had to remain stable in its placed configuration for at least 3 seconds after the XHAND1 retracted, without toppling or rolling away.

        \item Overall Trial Success: A trial was deemed successful if and only if all three criteria (Grasp, Transport, and Placement) were met.
    \end{enumerate}

    \textbf{Task 2: Dual-Arm Manipulation (Pick, Handoff, Place) with AgileX PiPERs}~~This task involved two AgileX PiPER robot arms collaboratively picking an object, handing it off from one arm to the other, and then placing it at a final target location. This sequence introduces challenges related to inter-arm coordination, synchronization, and stable object transfer.

    \textbf{1. Hardware Configuration.}~~

    \begin{enumerate}
        \item AgileX PiPER Robot Arms ($\times2$), the two PiPER arms were mounted on fixed bases, facing each other across a central workspace, 800 mm apart. Each PiPER arm was 6-DoF, equipped with an Agilex Pika Gripper.
    \end{enumerate}    
    
    \textbf{2. Task Protocol.}~~
    Similar to Task 1, objects were presented randomized within a $10\times 10 cm^2$ area with orientation randomized $\pm 15$ degrees around the vertical axis. Arm 1 was set as Giver, while Arm 2 was Receiver.

    A designated handoff procedure was followed. The giving arm (Arm 1) moved the object to a pre-defined handoff region (a specific area in the shared workspace). Arm 2 approached the object in Arm 1's gripper with a same grasp configuration, and then the receiving arm (Arm 2) placed the object at the table.

    Arm 1 always performed the initial pick and acted as the "giver"; Arm 2 always acted as the "receiver" and performed the final place. Arm motions were synchronized using an event-based system where Arm 2's approach was triggered by Arm 1 reaching the handoff pose, and Arm 1's release was triggered by a confirmation signal from Arm 2 (successful grasp confirmed by force sensors). The timing of gripper actions (Arm 2 closing, Arm 1 opening) was critical to prevent drops.

    An initial 6D pose estimate of the object was used by Arm 1 for the pick. During handoff, Arm 2 relied on the known pose of Arm 1's end-effector and visually track the object in Arm 1's gripper before Arm 2 grasped it. After handoff, the pose of the object was used by Arm 2 for planning the final placement.

    For each of the 15 objects, 30 pick-handoff-place trials were conducted. Scene and object reset procedures were followed between trials.

    \textbf{3. Success Criteria.}~~
    \begin{enumerate}
        \item Successful Pick (Arm 1): The object was securely grasped by Arm 1 and moved to the designated pre-handoff pose without slipping, dropping, or collision.
        \item Successful Handoff: Arm 2 securely grasped the object from Arm 1, arm 1 released the object only after Arm 2's grasp was confirmed via threshold met on the gripper force sensor (5 N). The object was not dropped during the transfer from Arm 1 to Arm 2 and the object was stable in Arm 2's gripper after Arm 1 retracted.
        \item Successful Place (Arm 2): The object was released by Arm 2 at the table, without slipping or being dropped.
        \item Overall Task Success: A trial was considered successful if and only if all three stages (Pick, Handoff, and Place) were completed successfully.
    \end{enumerate}
    \begin{figure}[htbp]
\centerline{
\includegraphics[width=1\columnwidth]{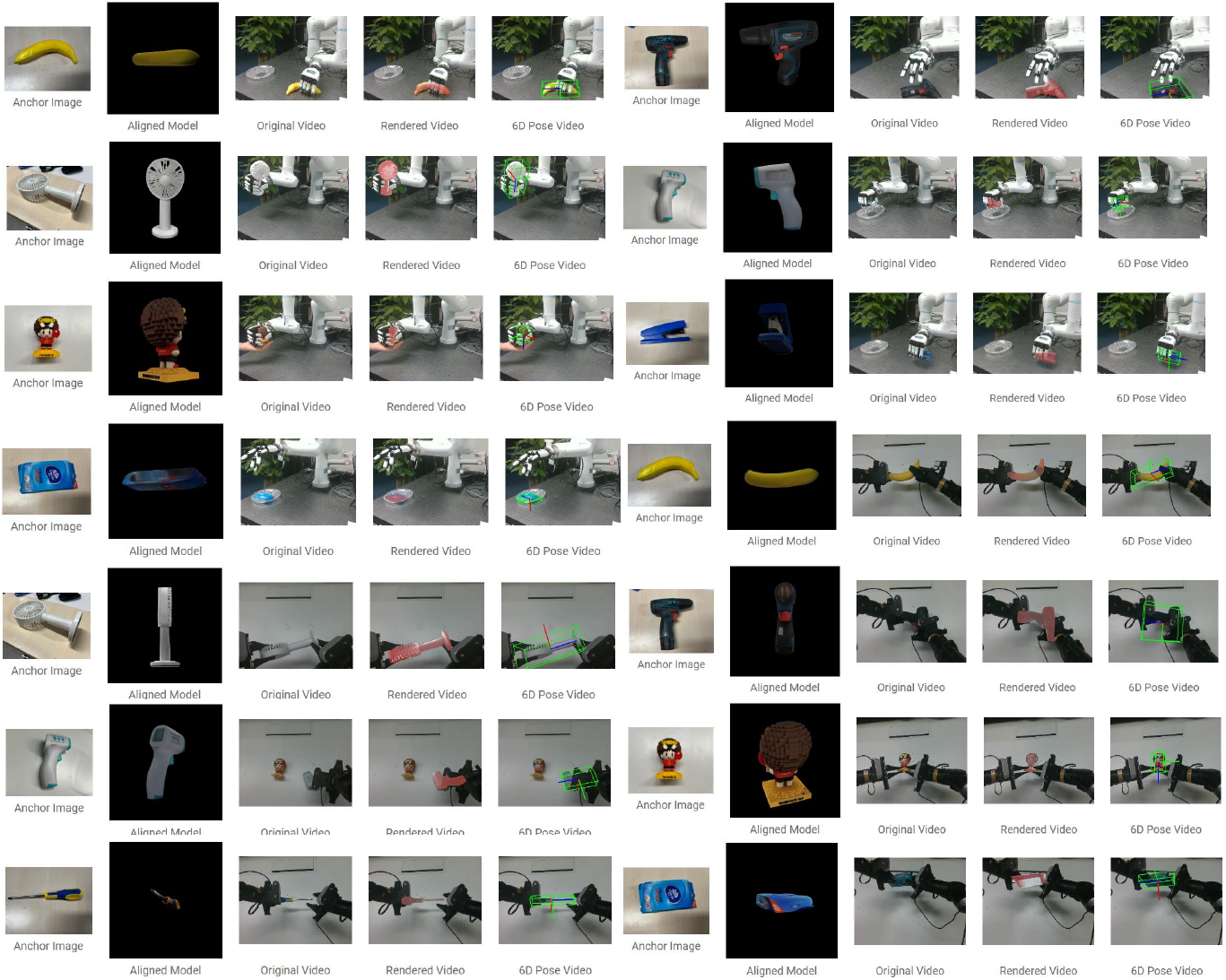}
}

\caption{
\textbf{Real-world Robot Manipulation Examples. }We recommend that readers visit the \href{https://gzwsama.github.io/OnePoseviaGen.github.io/}{webpage} to watch a dynamic demonstration video.
}

\label{fig:real-world-examples}
\end{figure}

\section{Pipeline for Constructing a Ground Truth Real-world Dataset}
\label{annotated_dataset}

\begin{figure}[htbp]
\centerline{
\includegraphics[width=1\columnwidth]{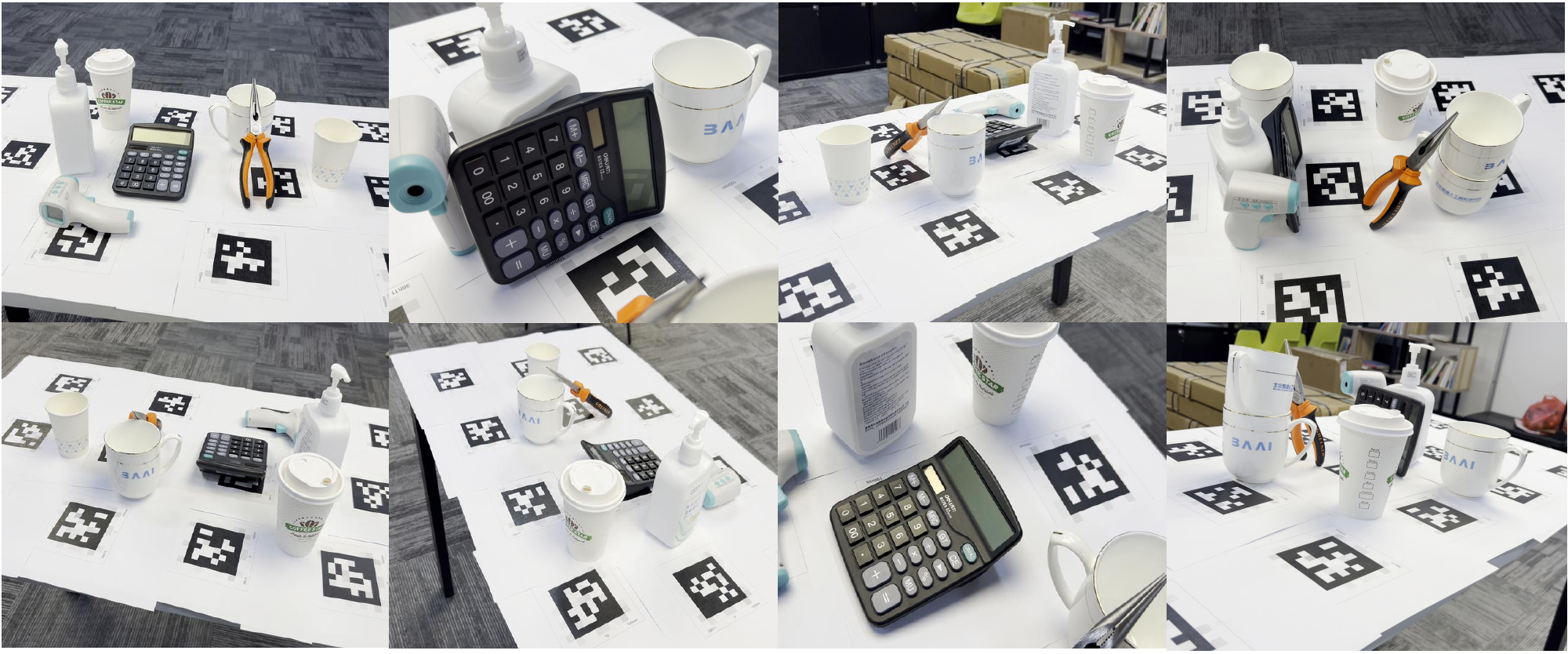}
}

\caption{
\textbf{Testing dataset examples. } The dataset we have constructed is showcased from various views. It can be seen that our testing dataset includes a range of object poses, occlusion relationships, and distances from the camera, demonstrating its richness and diversity.
}

\label{fig:test_dataset}
\end{figure}

\paragraph{Scene data acquisition and preprocessing.} In this paragraph, we describe the complete pipeline for scene data acquisition and preprocessing, which includes four key stages:
\begin{enumerate}
    \item \textbf{Deployment of AprilTags and Video Recording:} To facilitate accurate camera pose calculation and metric scale recovery, multiple high-contrast and easily detectable AprilTags~\cite{wang2016apriltag} are strategically placed throughout the target scene. As shown in Fig.~\ref{fig:test_dataset},these tags serve as known geometric reference points, significantly improving the robustness and accuracy of the reconstruction process. A monocular camera is then used to record a video sequence at a fixed frame rate, ensuring full coverage of the most of AprilTags from multiple viewpoints.

    \item \textbf{Feature Matching and Sparse Reconstruction Using COLMAP:} Structure-from-Motion (SfM)~\cite{schonberger2016sfm} processing is performed using COLMAP. First, local feature descriptors are extracted from each frame. These features are then matched across images, followed by geometric verification to remove outliers. Finally, an incremental SfM algorithm jointly optimizes both intrinsic and extrinsic camera parameters, resulting in a sparse 3D point cloud. This sparse reconstruction provides initial geometric priors for subsequent dense reconstruction.

    \item \textbf{Distortion Correction:} Due to radial and tangential distortions inherent in standard camera lenses, which can degrade reconstruction quality, it is essential to perform distortion correction. The raw images are undistorted and mapped to a distortion-free coordinate system. Simultaneously, the corresponding camera parameter files are updated to reflect this transformation, providing more accurate input for the 3D reconstruction pipeline.

    \item \textbf{Scale Estimation and Point Cloud Calibration Using XRSFM:} In purely vision-based SfM pipelines, the reconstructed scene is typically only defined up to a similarity transformation, lacking true metric scale. By leveraging the known physical dimensions of the AprilTags, the XRSFM~\cite{xrsfm} method is employed to compute the global scaling factor. This allows the sparse point cloud to be transformed from a similarity space into a Euclidean space with real-world units, enabling downstream tasks that require metric accuracy.
\end{enumerate}

\paragraph{3D Reconstruction and Object Mesh Extraction.}
After metric calibration of the point cloud, we utilize Gaussian Opacity Fields (GOF)~\cite{Yu2024GOF} to perform neural implicit field modeling of the scene. This method embeds sparse point cloud data into a Gaussian representation and models surface geometry through a learnable opacity field. Specifically, the point cloud is first initialized, and a set of spatially supported Gaussian attributes is constructed. The model is then trained using radiance optimization combined with geometry-aware training strategies, enabling the implicit representation of fine-grained surface details. Finally, a triangle mesh is extracted from the learned GOF representation via marching cubes or analogous voxelization techniques, yielding a high-quality surface reconstruction of the scene. The target object mesh is then manually segmented from the reconstructed scene mesh for further processing.

\paragraph{Coordinate alignment and camera/object pose estimation.} This paragraph introduces the complete workflow for coordinate alignment and 6D pose estimation, including initial camera pose initialization, camera poses calculation, and per-frame object pose computation:
\begin{enumerate}
    \item \textbf{Aligning Object and Scene Meshes for Initial Camera Pose Estimation:} The object mesh is rigidly aligned with the scene mesh using ICP (Iterative Closest Point) registration algorithms. This provides the object’s absolute position within the scene. Combined with AprilTag location information, this enables derivation of the initial camera pose $\mathbf{T}_{C_0}^{W}$ (rotation and translation matrices in the world coordinate system) for the first frame.

    \item \textbf{Estimating Per-Frame Camera Poses Using COLMAP Features:} Based on the camera poses $\mathbf{T}_{C_i}^{W}$ output by COLMAP (relative to the sparse point cloud), and incorporating the previously estimated metric scale, all camera poses are transformed into the unified world coordinate system. This step establishes a reliable camera trajectory that supports accurate object position computation.

    \item \textbf{Computing Object Poses Across Frames:} Assuming the object remains static in the scene, its pose in the first frame can be transformed into any other frame's camera coordinate system through a rigid transformation chain. Specifically, given the object's pose $\mathbf{T}_{O_1}^{W}$ in the world coordinate system (where subscript $O_1$ denotes the object in the first frame and superscript $W$ denotes the world coordinate system), and the camera poses $\mathbf{T}_{C_i}^{W}$ for each frame $i$, we can compute the object's pose $\mathbf{T}_{O_i}^{C_i}$ in the camera coordinate system of frame $i$ using the following transformation:

    \begin{equation}
        \mathbf{T}_{O_i}^{C_i} = (\mathbf{T}_{C_i}^{W})^{-1} \cdot \mathbf{T}_{O_1}^{W}
    \end{equation}

    Here, $\mathbf{T}_{C_i}^{W}$ represents the transformation matrix from the world coordinate system to the camera coordinate system of frame $i$, and $(\mathbf{T}_{C_i}^{W})^{-1}$ is its inverse, which transforms points from the camera coordinate system back to the world coordinate system.

    The 6-DoF object pose in each image frame consists of three rotational parameters ($\mathbf{R}_{O_i}^{C_i}$) and three translational parameters ($\mathbf{t}_{O_i}^{C_i}$). These can be extracted from the transformation matrix $\mathbf{T}_{O_i}^{C_i}$ as follows:

    \begin{equation}
        \mathbf{T}_{O_i}^{C_i} = 
        \begin{bmatrix}
            \mathbf{R}_{O_i}^{C_i} & \mathbf{t}_{O_i}^{C_i} \\
            0 & 1
        \end{bmatrix}
    \end{equation}
\end{enumerate}

\paragraph{Object mask generation and visualization details.} This paragraph introduces the methodology for generating accurate object masks and visualizing key details, which includes two main steps:
\begin{enumerate}
    \item \textbf{Rendering Object Masks Using 3D Models and Camera Poses:} Given the known 3D mesh model of the object and the current frame's camera parameters (intrinsic + extrinsic), a binary object mask is rendered onto the image plane. This forward projection process generates a foreground-background segmentation that supports downstream tasks such as pose estimation and visibility analysis.

    \item \textbf{Extracting Visible Region Masks Using Segmentation Methods (e.g., SAM 2):} To further improve mask accuracy—especially under occlusion, a pre-trained instance segmentation method such as SAM 2~\cite{kirillov2023segany} is applied to the input images. This generates a visible region mask highlighting only the observable parts of the object, which can be used during training or evaluation to exclude occluded or invisible regions and improve robustness.
\end{enumerate}
\begin{figure}[htbp]
\centerline{
\includegraphics[width=0.8\columnwidth]{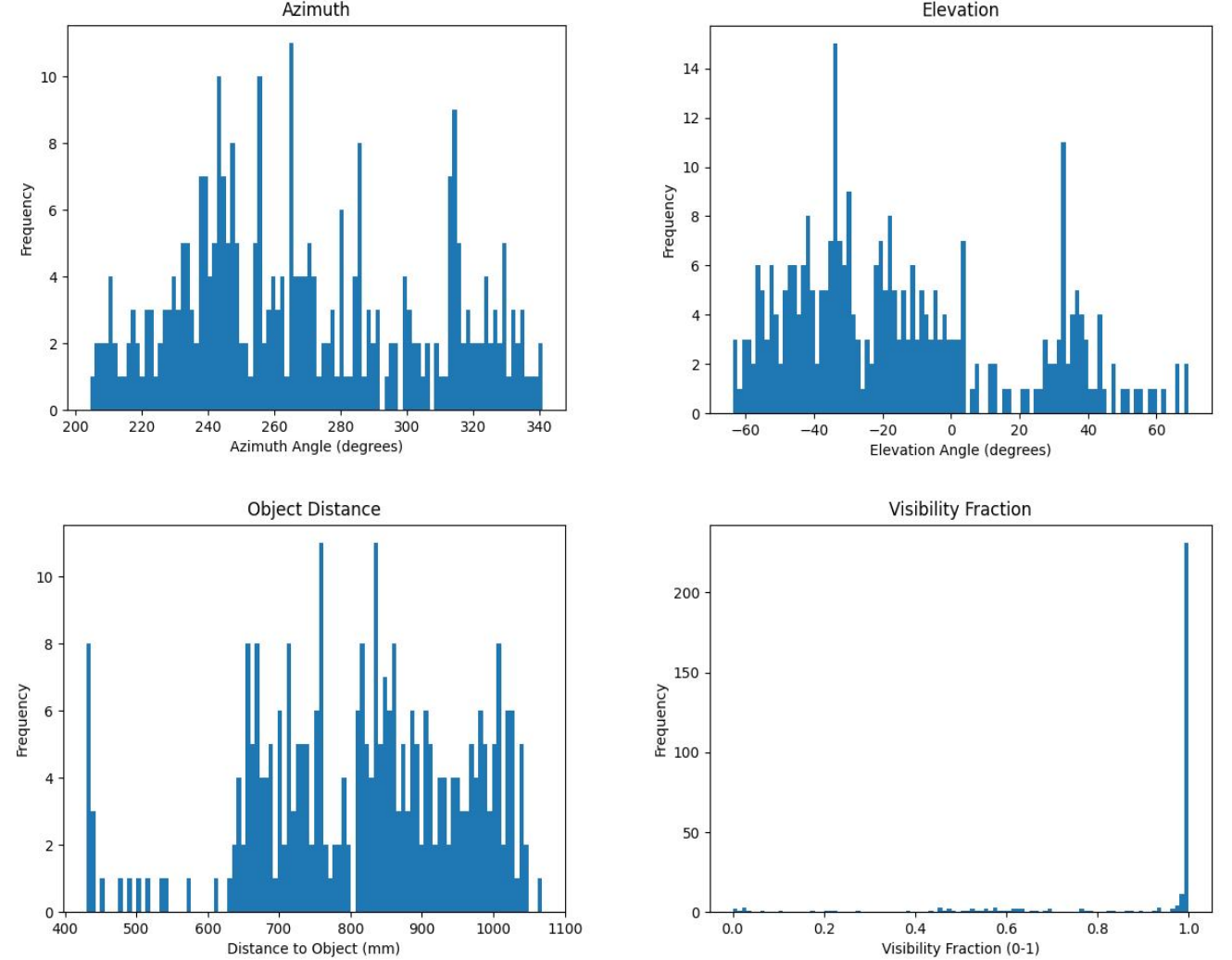}
}

\caption{
\textbf{Distribution Analysis of Testing Dataset. } Top-left: Distribution of Azimuth angles for objects within the generated dataset;
Top-right: Distribution of Elevation angles for objects;
Bottom-left: Distribution of object distances from the camera;
Bottom-right: Distribution of the proportion of object area visibility;
These plots demonstrate that the constructed testing dataset encompasses a rich diversity of poses and occlusion scenarios, underpinning its comprehensive variability and realism.
}

\label{fig:test_dataset_distribution}
\end{figure}

Finally, we constructed the following test dataset. The dataset visualization is shown in Fig.~\ref{fig:test_dataset}, and its distribution is illustrated in Fig.~\ref{fig:test_dataset_distribution}. These figures demonstrate that the dataset exhibits diverse object positions and viewing angles, making it well-suited for comprehensively evaluating the performance of the trained models.

%===============================================================================

%===============================================================================
% \clearpage
% \bibliography{appendex}
% \end{document}

\end{document}